\documentclass{article}
\usepackage{ijcai19}

\usepackage{times}
\usepackage{soul}
\usepackage{url}
\usepackage[hidelinks]{hyperref}
\usepackage[utf8]{inputenc}
\usepackage[small]{caption}
\usepackage{graphicx}
\usepackage{amsmath}
\usepackage{booktabs}
\usepackage{amsfonts}
\usepackage{amsmath}
\usepackage{graphicx}
\usepackage{caption}
\usepackage{subcaption}
\usepackage{algorithm}
\usepackage{algpseudocode}
\usepackage{multirow}
\usepackage{array}
\usepackage{float}
\usepackage[OT1]{fontenc}

\title{Disentangled Makeup Transfer with Generative Adversarial Network}

\author{
Honglun Zhang$^{1,2}$\and
Jidong Tian$^{1,2}$\and
Wenqing Chen$^{1,2}$\and
Hao He$^{2}$\and
Yaohui Jin$^{1,2}$\footnote{Contact Author}\\
\affiliations
$^1$MoE Key Lab of Artificial Intelligence, AI Institute, Shanghai Jiao Tong University\\
$^2$State Key Lab of Advanced Optical Communication System and Network
\emails
\{jinyh\}@sjtu.edu.cn
}

\begin{document}

\maketitle

\begin{abstract}
Facial makeup transfer is a widely-used technology that aims to transfer the makeup style from a reference face image to a non-makeup face. Existing literature leverage the adversarial loss so that the generated faces are of high quality and realistic as real ones, but are only able to produce fixed outputs. Inspired by recent advances in disentangled representation, in this paper we propose \textbf{DMT}~(Disentangled Makeup Transfer), a unified generative adversarial network to achieve different scenarios of makeup transfer. Our model contains an identity encoder as well as a makeup encoder to disentangle the personal identity and the makeup style for arbitrary face images. Based on the outputs of the two encoders, a decoder is employed to reconstruct the original faces. We also apply a discriminator to distinguish real faces from fake ones. As a result, our model can not only transfer the makeup styles from one or more reference face images to a non-makeup face with controllable strength, but also produce various outputs with styles sampled from a prior distribution. Extensive experiments demonstrate that our model is superior to existing literature by generating high-quality results for different scenarios of makeup transfer.

\end{abstract}

\section{Introduction}

Makeup is a widely-used skill to improve one's facial appearance but it is never easy to become a professional makeup artist as there are so many different cosmetic products and tools diverse in brands, categories and usages. As a result, it has been increasingly popular to try different makeup styles on photos or short videos with virtual makeup software. \textit{Facial Makeup Transfer}~\cite{DBLP:conf/pg/TongTBX07} provides an effective solution to this by naturally transferring the makeup style from a well-suited reference face image to a non-makeup face, which can be utilized in a wide range of applications like photograph, video, entertainment and fashion.

Rather than traditional image processing methods~\cite{DBLP:conf/cvpr/GuoS09,DBLP:conf/cvpr/LiZL15,DBLP:conf/pg/TongTBX07} like image gradient editing and physics-based manipulation, recent literature on facial makeup transfer mostly employ deep neural networks~\cite{DBLP:journals/pami/BengioCV13} to learn the mapping from non-makeup face images to makeup ones, and leverage the adversarial loss of GAN~(\textit{Generative Adversarial Network})~\cite{DBLP:conf/nips/GoodfellowPMXWOCB14} to generate realistic fake images. In order to accurately capture the makeup style, several methods~\cite{DBLP:conf/ijcai/LiuOQWC16,DBLP:conf/mm/LiQDLYZL18} have been proposed to evaluate the differences between the generated face and the reference face on crucial cosmetic components like foundation, eyebrow, eye shadow and lipstick. 

\begin{figure*}[tb!]
\centering
\includegraphics[width=0.9\textwidth]{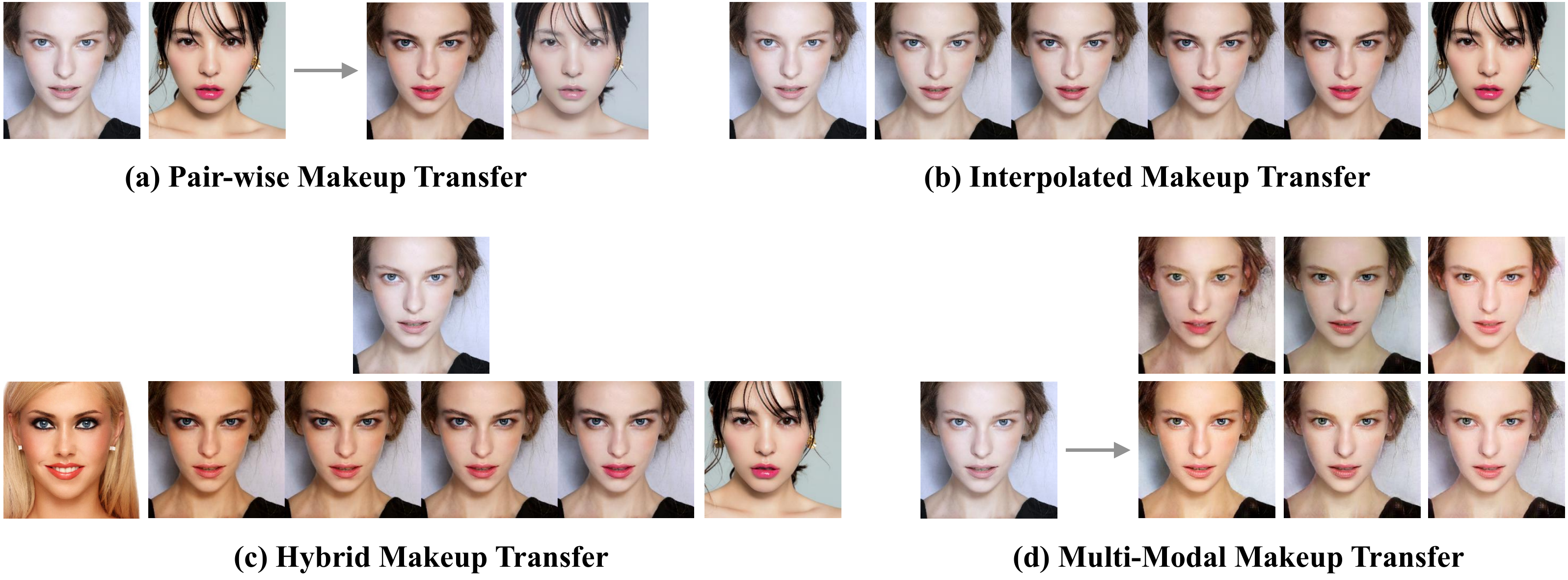}
\caption{Different scenarios of makeup transfer. Most related researches only focus on the pair-wise makeup transfer. In contrast, our model can achieve all of them.}\label{zhang1}
\end{figure*}

However, existing approaches mainly focus on makeup transfer between two face images and are only able to produce fixed outputs, which is denoted by \textit{pair-wise makeup transfer} as Fig.\ref{zhang1} illustrates. In fact, there are several other scenarios of makeup transfer, such as controlling the strength of makeup style~(\textit{interpolated makeup transfer}), blending the makeup styles of two or more reference images~(\textit{hybrid makeup transfer}) and producing various outputs based on a single non-makeup face without any reference images~(\textit{multi-modal makeup transfer}). To the best of our knowledge, those different scenarios have not been researched much yet and cannot be well achieved by existing literature.

In this paper, we propose \textbf{DMT}~(Disentangled Makeup Transfer), a unified generative adversarial network to achieve different scenarios of makeup transfer. Inspired by recent advances in \textit{Disentangled Representation}~\cite{DBLP:conf/eccv/HuangLBK18,DBLP:conf/cvpr/MaSGGSF18,DBLP:conf/eccv/LeeTHSY18}, our model utilizes an identity encoder as well as a makeup encoder to disentangle the \textit{personal identity} and the \textit{makeup style} for arbitrary face images. Based on the outputs of the two encoders, we further employ a decoder to reconstruct the original faces. We also apply a discriminator to distinguish real face images from fake ones. Thanks to such a disentangled architecture, our model can not only transfer the makeup styles from one or more reference face images to a non-makeup face with controllable strength, but also produce various outputs with styles sampled from a prior distribution. Furthermore, we leverage the attention mask~\cite{DBLP:conf/eccv/ChenXYT18,DBLP:conf/nips/MejjatiRTCK18,DBLP:journals/corr/abs-1806-06195,DBLP:conf/eccv/ZhangKSC18} to refine the transfer results so that the makeup-unrelated content is well preserved. We perform extensive experiments on a dataset that contains both makeup and non-makeup face images~\cite{DBLP:conf/mm/LiQDLYZL18}. Both qualitative and quantitative results demonstrate that our model is superior to existing literature by generating high-quality faces for different makeup transfer scenarios.

Our contributions are summarized as follows.

\begin{itemize}
    \item We propose DMT, a unified model to achieve different scenarios of makeup transfer. To the best of our knowledge, we are the first to integrate disentangled representation to solve the task of facial makeup transfer.
    \item With such a disentangled architecture, our model is able to conduct different scenarios of makeup transfer, including pair-wise, interpolated, hybrid and multi-modal, which cannot be achieved by related researches.
    \item Extensive experiments demonstrate the superiority of our model against state-of-the-arts, both qualitatively and quantitatively.
\end{itemize}

\section{Related Work}

\noindent\textbf{Generative Adversarial Network}~(GAN)~\cite{DBLP:conf/nips/GoodfellowPMXWOCB14} is a powerful method for training generative models of complex data and has been proved effective in a wide range of computer vision tasks, including image generation~\cite{DBLP:journals/corr/RadfordMC15,DBLP:conf/iccv/ZhangXL17,DBLP:conf/nips/GulrajaniAADC17}, image-to-image translation~\cite{DBLP:conf/cvpr/IsolaZZE17,DBLP:journals/corr/ZhuPIE17,DBLP:journals/corr/abs-1711-09020}, inpainting~\cite{DBLP:conf/eccv/LiuRSWTC18}, super-resolution~\cite{DBLP:conf/cvpr/LedigTHCCAATTWS17} and so on. In this paper, we leverage the adversarial loss of GAN to generate realistic faces that are indistinguishable from real ones.\vspace{1mm}

\noindent\textbf{Facial Makeup Transfer} aims to transfer the makeup style from a reference face image to a non-makeup face. Traditional methods include \cite{DBLP:conf/cvpr/GuoS09} and \cite{DBLP:conf/cvpr/LiZL15}, which decompose face images into several layers and conduct makeup transfer within each layer. \cite{DBLP:conf/ijcai/LiuOQWC16} proposes an optimization-based deep localized makeup transfer network and applies different transfer methods for different cosmetic components. In contrast, \cite{DBLP:conf/mm/LiQDLYZL18} trains a learning-based model with dual inputs and outputs to achieve pair-wise makeup transfer and only requires a forward pass for inference. Other related topics include \textit{Unpaired Image-to-Image Translation}~\cite{DBLP:journals/corr/ZhuPIE17,DBLP:conf/icml/KimCKLK17,DBLP:conf/iccv/YiZTG17}, where images of two domains are translated bi-directionally, and \textit{Style Transfer}~\cite{DBLP:journals/corr/GatysEB15a,DBLP:conf/eccv/JohnsonAF16}, where a transfer image is synthesized based on a content image and a style image. However, an image-to-image translation model trained on makeup and non-makeup faces can only learn domain-level mappings, thus producing fixed output for a certain non-makeup face regardless of the reference image. Style transfer models can be used to conduct makeup transfer by treating the makeup and non-makeup faces as the style image and the content image respectively, but can only learn global features of the whole images and fails to focus on crucial cosmetic components.\vspace{1mm}

\noindent\textbf{Disentangled Representation} means decomposing the original input into several independent hidden codes so that the features of different components can be better learned. \cite{DBLP:conf/eccv/HuangLBK18,DBLP:conf/eccv/LeeTHSY18} disentangle images into domain-invariant content codes and domain-specific style codes to obtain multi-modal outputs for unpaired image-to-image translation tasks. \cite{DBLP:conf/cvpr/MaSGGSF18} introduces a disentangled representation of three components, the foreground, the background and the body pose, to manipulate images of pedestrians and fashion models. In this paper, we propose to disentangle an arbitrary face image into two independent components, the personal identity and the makeup style.\vspace{1mm}

\noindent\textbf{Attention Mask} is an effective mechanism widely-used in image-to-image translation~\cite{DBLP:conf/eccv/ChenXYT18,DBLP:conf/nips/MejjatiRTCK18,DBLP:journals/corr/abs-1806-06195} and image editing~\cite{DBLP:conf/eccv/ZhangKSC18} tasks, which learns to localize the interested region and preserve the unrelated content. In this paper, we employ the attention mask in our model so that the makeup-unrelated region including the hair, the clothing and the background keeps unchanged after transfer.

\begin{figure}[tb!]
\centering
\includegraphics[width=0.48\textwidth]{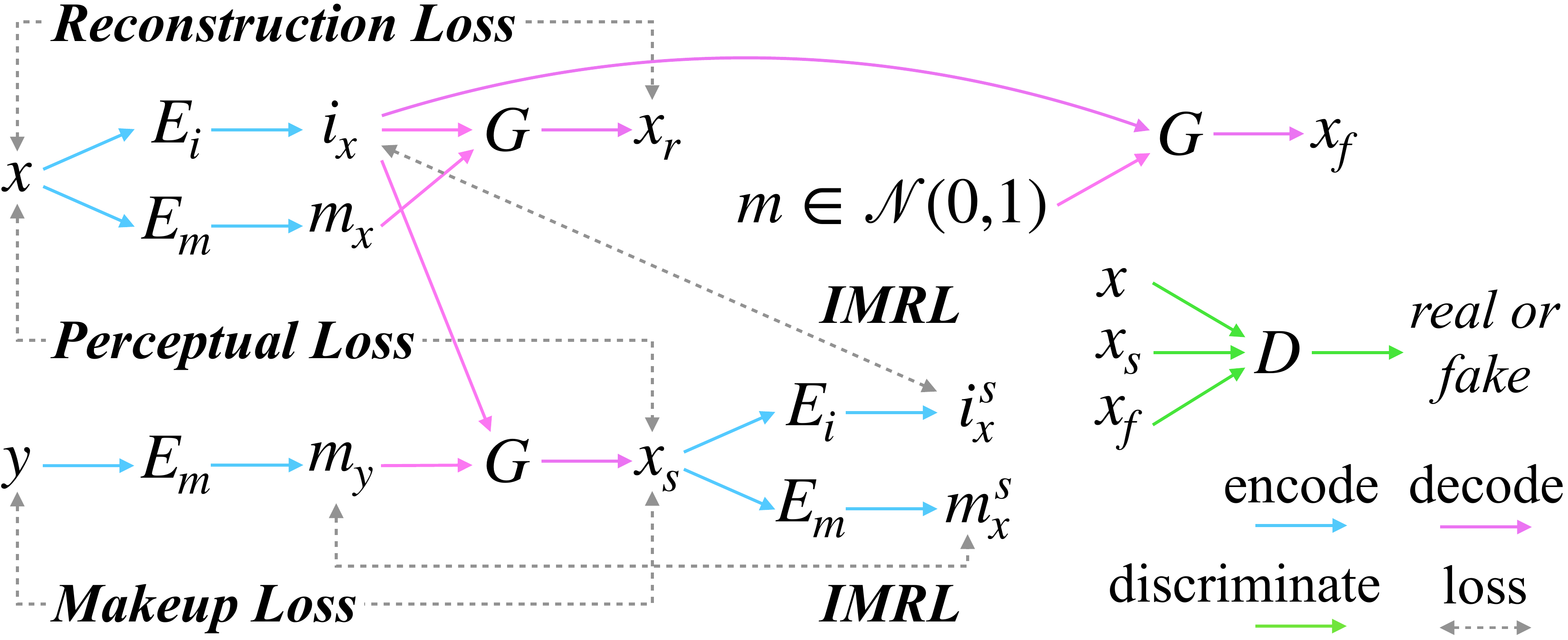}
\caption{The disentangled architecture of DMT, which contains four modules in all, the Identity Encoder $E_i$, the Makeup Encoder $E_m$, the Decoder $G$ and the Discriminator $D$.}\label{zhang2}
\end{figure}

\section{Methodology}

\subsection{Disentangled Makeup Transfer}

For a given face image $x\in[0,1]^{H\times W\times 3}$, which can be either a non-makeup face or a makeup one with arbitrary style, we propose to disentangle it into two components that are independent of each other, the \textit{personal identity} and the \textit{makeup style}. As a result, facial makeup transfer can be achieved by combining the same personal identity with different makeup styles, just like a person wearing different clothes.

Based on the above assumption, we propose \textbf{DMT}~(Disentangled Makeup Transfer), a unified and flexible generative adversarial network to conduct different scenarios of makeup transfer. As Fig.\ref{zhang2} shows, our model contains four modules in all, the \textbf{Identity Encoder} $E_i$, the \textbf{Makeup Encoder} $E_m$, the \textbf{Decoder} $G$ and the \textbf{Discriminator} $D$. For the given face image $x$, we obtain the corresponding \textbf{Identity Code} $i_x$ and \textbf{Makeup Code} $m_x$ with $E_i$ and $E_m$ as follows.
\begin{align*}
&i_x=E_i(x)\tag{$1$}\label{eq:1}\\
&m_x=E_m(x)\tag{$2$}\label{eq:2}
\end{align*}

We suppose that $m_x$ captures the makeup style of $x$ including crucial cosmetic components like foundation, eyebrow, eye shadow and lipstick, whereas $i_x$ conveys the information of other makeup-unrelated content such as personal identity, clothing and background. Furthermore, $i_x$ and $m_x$ should be independent of each other as they describe different features of $x$, which satisfies the definition of disentangled representation. Based on $i_x$ and $m_x$, we leverage $G$ to obtain the reconstructed image $x_r$ without loss of information after encoding and decoding, which can be regulated by the following \textbf{Reconstruction Loss}.
\begin{align*}
&x_r=G(i_x,m_x)\tag{$3$}\label{eq:3}\\
&\mathcal{L}_{rec}^G=\mathbb{E}_x[\left\|x-x_r\right\|_1]\tag{$4$}\label{eq:4}
\end{align*}
where $\left\|\cdot\right\|_1$ is the $\mathcal{L}_1$ norm used to calculate the absolute difference between $x$ and $x_r$.

\subsection{Pair-wise Makeup Transfer}

Pair-wise makeup transfer aims to swap the makeup styles of two face images, thus producing an after-makeup face and an anti-makeup face as Fig.\ref{zhang1} shows. Given another face image $y\in[0,1]^{H\times W\times 3}$, we apply $E_m$ again to obtain the corresponding makeup code $m_y$ as Fig.\ref{zhang2} shows.
\begin{equation}\tag{$5$}\label{eq:5}
m_y=E_m(y)
\end{equation}

Based on $i_x$ and $m_y$, we obtain the transfer result $x_s$ as follows, which is supposed to preserve the personal identity of $x$ and synthesize the makeup style of $y$.
\begin{equation}\tag{$6$}\label{eq:6}
x_s=G(i_x,m_y)
\end{equation}

It should be noted that both $x$ and $y$ can be either makeup or non-makeup faces, thus leading to four different cases of pair-wise makeup transfer as Table \ref{tab:1} shows, which well cover the objectives investigated in most related researches. For training, we randomly set $x$ and $y$ as makeup or non-makeup images with equal probabilities, which helps our model learn to handle different cases.

\begin{table}[tb!]
\centering
\begin{tabular}{c c c c} \hline
$x$ & $y$ & $x_s$ & Objective\\ \hline
- & - & - & - \\
- & \checkmark & \checkmark & add makeup \\
\checkmark & - & - & remove makeup \\
\checkmark & \checkmark & \checkmark & swap makeup \\ \hline
\end{tabular}
\caption{Four different cases of pair-wise makeup transfer, where - means non-makeup and $\checkmark$ means makeup.}\label{tab:1}
\end{table}

As for personal identity preservation, it is improper to directly compare $x$ and $x_s$ in the raw pixel-level. Instead, we utilize a VGG-16~\cite{DBLP:journals/corr/SimonyanZ14a} model pre-trained on the ImageNet dataset~\cite{DBLP:journals/ijcv/RussakovskyDSKS15} to compare their activations in a certain hidden layer, as it has been proved that deep neural networks are effective in extracting high-level features~\cite{DBLP:journals/corr/GatysEB15a}. In order to preserve the personal identity of $x$, we employ the following \textbf{Perceptual Loss} to measure the difference between $x$ and $x_s$ in the $l$-th layer of VGG-16.
\begin{equation}\tag{$7$}\label{eq:7}
\mathcal{L}_{per}^G=\mathbb{E}_{x,y}[\left\|A_l(x)-A_l(x_s)\right\|_2]
\end{equation}
where $\left\|\cdot\right\|_2$ is the $\mathcal{L}_2$ norm and $A_l(\cdot)$ denotes the output of the $l$-th layer. By minimizing the above perceptual loss, we can ensure that the original high-level features of $x$ is well preserved in $x_s$.

\begin{figure}[tb!]
\centering
\includegraphics[width=0.45\textwidth]{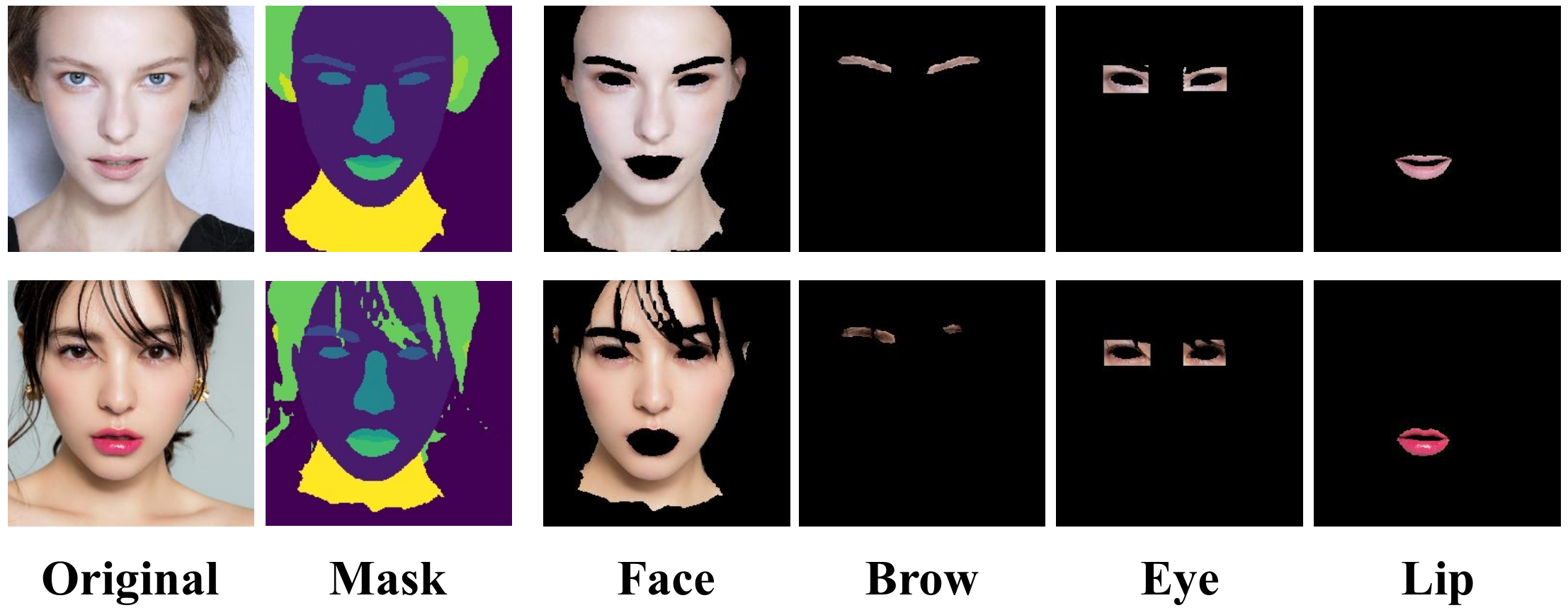}
\caption{Examples of parsing masks and cosmetic regions.}\label{zhang3}
\end{figure}

\begin{figure*}[tb!]
\centering
\includegraphics[width=0.9\textwidth]{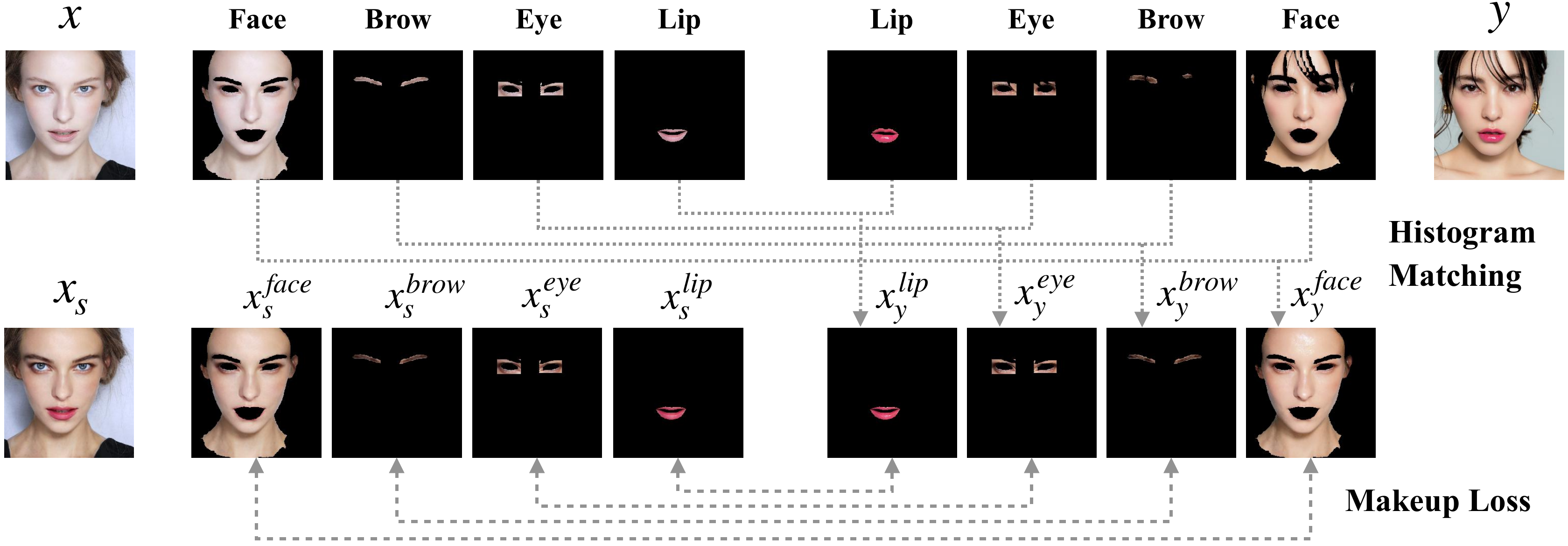}
\caption{Calculation of the makeup loss. We first perform histogram matching on different cosmetic regions of $x$ and $y$ to produce a ground truth $x_y$, which shares the same color distribution as $y$ on each region and preserves the shape information of $x$, then calculate the makeup loss between $x_s$ and $x_y$ on each cosmetic region.}\label{zhang4}
\end{figure*}

Another challenge is how to evaluate the instance-level consistency of $y$ and $x_s$ in makeup style. Here we leverage the \textbf{Makeup Loss} proposed by \cite{DBLP:conf/mm/LiQDLYZL18}. As Fig.\ref{zhang3} shows, we obtain the parsing mask for each face image, which consists of $14$ semantic parts, \textit{background}, \textit{face}, \textit{left\,/\,right eyebrow}, \textit{left\,/\,right eye}, \textit{nose}, \textit{upper\,/\,lower lip}, \textit{mouth}, \textit{hair}, \textit{left\,/\,right ear}, \textit{neck}, and can be achieved by training a semantic segmentation model~\cite{DBLP:conf/cvpr/ZhaoSQWJ17,DBLP:conf/eccv/YuWPGYS18} on face parsing datasets~\cite{DBLP:conf/cvpr/SmithZBLY13,CelebAMask-HQ}. Based on the parsing mask, we extract the following four regions to cover crucial cosmetic components for each face image.

\begin{itemize}
    \item \textbf{Face} covers the foundation, including \textit{face}, \textit{nose}, \textit{left\,/\,right ear}, \textit{neck}.
    \item \textbf{Brow} covers the eyebrow, including \textit{left\,/\,right eyebrow}.
    \item \textbf{Eye} covers the eye shadow. We extract two rectangle regions enclosing the eyes and exclude overlapping content of \textit{hair}, \textit{left\,/\,right eye}, \textit{left\,/\,right eyebrow}.
    \item \textbf{Lip} covers the lipstick, including \textit{upper\,/\,lower lip}. 
\end{itemize}

As Fig.\ref{zhang3} shows, the makeup style of each cosmetic region mainly depends on the color distribution. For example, adding lipstick for the non-makeup face in Fig.\ref{zhang3} can be simply achieved by replacing the lip color with that of the makeup face. Therefore, the transfer result $x_s$ is supposed to share similar color distribution with $y$ on each cosmetic region. To meet this requirement, we first perform histogram matching on different regions of $x$ and $y$ to produce a ground truth $x_y$ as Fig.\ref{zhang4} shows, which shares the same color distribution as $y$ on each region and preserves the shape information of $x$. Then we calculate the makeup loss on different cosmetic regions of $x_s$ and $x_y$ with the $\mathcal{L}_2$ norm as follows.
\begin{equation}\tag{$8$}\label{eq:8}
\mathcal{L}_{mak}^G=\sum_{c\in\mathbb{C}}\lambda_{c}\mathbb{E}_{x,y}[\left\|x_s^c-x_y^c\right\|_2]
\end{equation}
where $\mathbb{C}=\{face,brow,eye,lip\}$, $x_s^c$ and $x_y^c$ denote the corresponding cosmetic regions of $x_s$ and $x_y$ for $c$, $\lambda_{face}$, $\lambda_{brow}$, $\lambda_{eye}$, $\lambda_{lip}$ are the weights to combine different loss terms.

Based on the perceptual loss and the makeup loss, the transfer result $x_s$ generated by $G$ not only preserves the personal identity of $x$ but also satisfies the makeup style of $y$. As Fig.\ref{zhang2} shows, we apply the encoders again on $x_s$ to obtain the corresponding identity code $i_x^s$ and makeup code $m_x^s$.
\begin{align*}
&i_x^s=E_i(x_s)\tag{$9$}\label{eq:9}\\
&m_x^s=E_m(x_s)\tag{$10$}\label{eq:10}
\end{align*}

To ensure the one-to-one mappings between face images and identity\,/\,makeup codes, we employ the following \textbf{Identity Makeup Reconstruction Loss}~(abbreviated as \textit{IMRL} in Fig.\ref{zhang2}) so that the disentangled representation keeps unchanged after decoding and encoding.
\begin{equation}\tag{$11$}\label{eq:11}
\mathcal{L}_{imr}^G=\lambda_{i}\mathbb{E}_{x,y}[\left\|i_x-i_x^s\right\|_1]+\lambda_{m}\mathbb{E}_{x,y}[\left\|m_y-m_x^s\right\|_1]
\end{equation}
where $\lambda_i$ and $\lambda_m$ are the weights of the identity term and the makeup term.

\subsection{Interpolated Makeup Transfer}

Interpolated makeup transfer is a general extension of pair-wise makeup transfer as it aims to control the strength of makeup style. Based on the disentangled representation discussed in previous sections, we can easily achieve this by combining the makeup styles of $x$ and $y$ with a controlling parameter $\alpha\in[0,1]$. As $\alpha$ increases from $0$ to $1$, the makeup style of the transfer result $x_s$ transits from $x$ to $y$ accordingly.
\begin{equation}\tag{$12$}\label{eq:12}
x_s=G(i_x,(1-\alpha)m_x+\alpha m_y)
\end{equation}

\subsection{Hybrid Makeup Transfer}

We can also achieve hybrid makeup transfer by blending multiple makeup styles. Given $K$ reference images $y^k$, we obtain their makeup codes $m_y^k$ and perform hybrid makeup transfer with controlling weights $\alpha^k,\sum_{k=1}^{K}{\alpha^k}=1$ as follows.
\begin{equation}\tag{$13$}\label{eq:13}
x_s=G(i_x,\sum_{k=1}^{K}{\alpha^k m_y^k})
\end{equation}

\subsection{Multi-Modal Makeup Transfer}

Multi-modal makeup transfer aims to produce various outputs based on a single non-makeup face without any reference images. As Fig.\ref{zhang2} shows, we randomly sample the makeup style $m$ from a prior distribution like the standard normal distribution $\mathcal{N}(0,1)$ and obtain the corresponding decoded result $x_f$.
\begin{equation}\tag{$14$}\label{eq:14}
x_f=G(i_x,m),m\in\mathcal{N}(0,1)
\end{equation}

\begin{figure*}[tb!]
\centering
\includegraphics[width=0.9\textwidth]{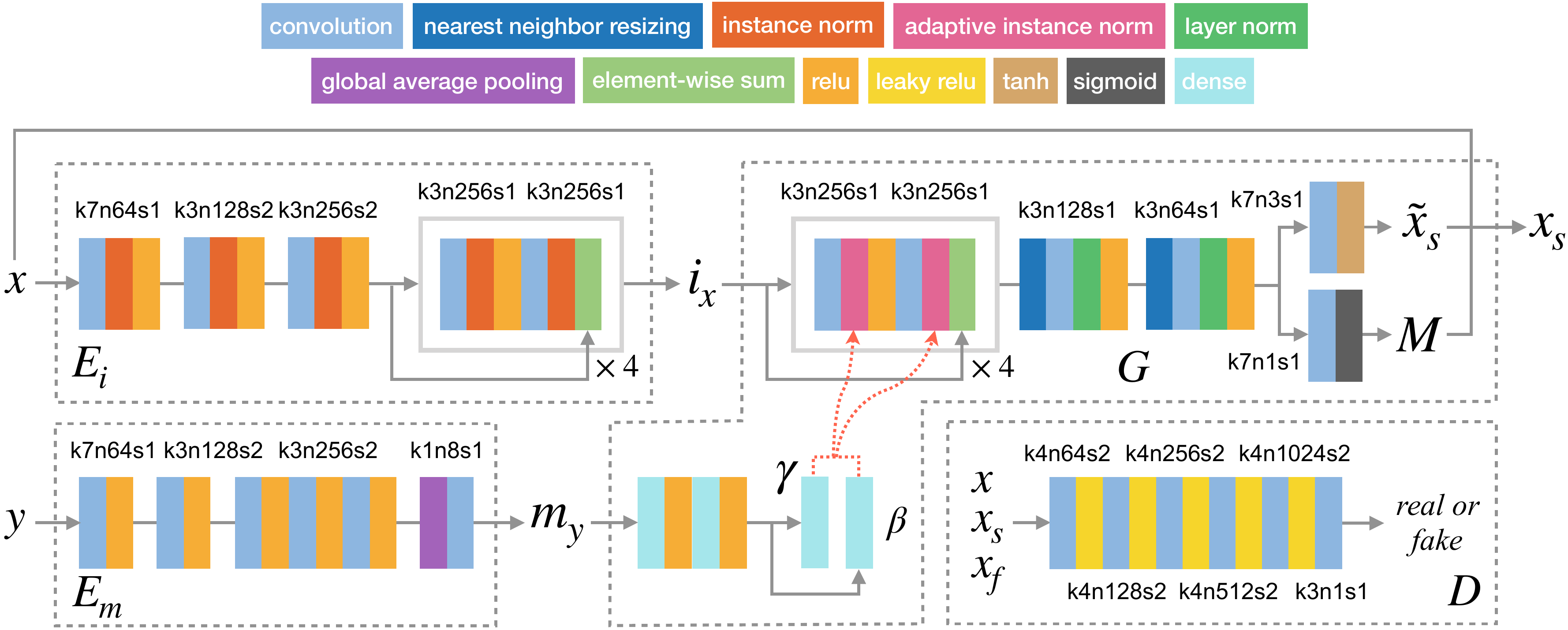}
\caption{Detailed structures of $E_i$, $E_m$, $G$ and $D$, where blocks of different colors denote different types of neural layers.}\label{zhang5}
\end{figure*}

As a result, $x_f$ only depends on $x$ and the random style $m$, and multi-modal makeup transfer can be achieved by sampling multiple styles to generate different outputs.

\subsection{Attention Mask}

We leverage the attention mask~\cite{DBLP:conf/eccv/ChenXYT18,DBLP:conf/nips/MejjatiRTCK18,DBLP:journals/corr/abs-1806-06195,DBLP:conf/eccv/ZhangKSC18} widely-used in image-to-image translation tasks to protect the makeup-unrelated content from being altered. Fig.\ref{zhang5} illustrates the network structure of DMT in details, where the model is utilized to conduct pair-wise makeup transfer between $x$ and $y$. Apart from generating the face image $\tilde{x}_s$, $G$ also learns to produce an attention mask $M\in[0,1]^{H\times W}$ to localize the makeup-related region, where higher values mean stronger relation. Based on the above definition of $M$, we obtain the refined result by selectively extracting the related content from $\tilde{x}_s$ and copying the rest from the original face $x$.
\begin{equation}\tag{$15$}\label{eq:15}
x_s=M\odot\tilde{x}_s + (1-M)\odot x
\end{equation}
where $\odot$ denotes element-wise multiplication and $1-M$ means inverting the mask to get the unrelated region.

\begin{figure}[tb!]
\centering
\includegraphics[width=0.25\textwidth]{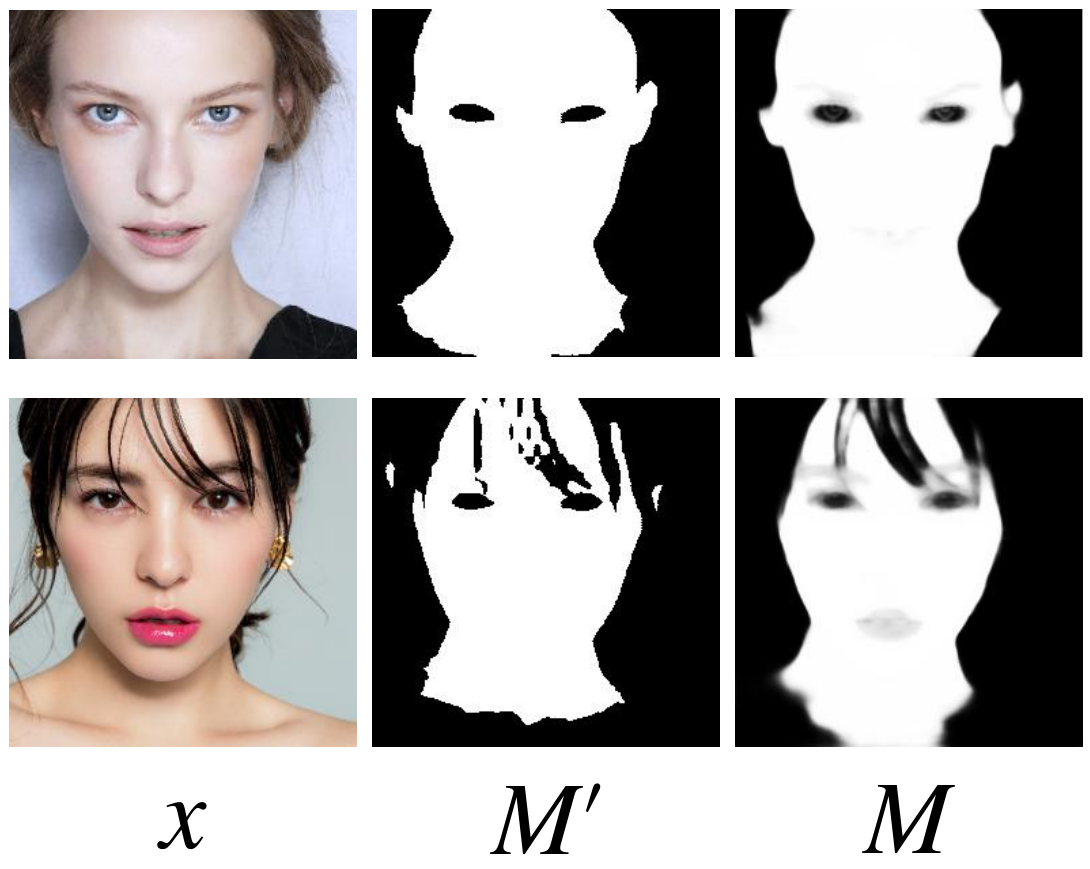}
\caption{Examples of makeup-related region ${M}'$ and the generated attention mask $M$.}\label{zhang6}
\end{figure}

As the parsing mask of each face image is available, we manually obtain the makeup-related region ${M}'$ as Fig.\ref{zhang6} shows by excluding \textit{background}, \textit{left\,/\,right eye} and \textit{hair} from the parsing mask, which can serve as the ground truth for $M$ by applying the \textbf{Attention Loss} as follows.
\begin{equation}\tag{$16$}\label{eq:16}
\mathcal{L}_{a}^G=\mathbb{E}_x[\left\|M-{M}'\right\|_1]
\end{equation}

\subsection{Other Loss Functions}

In this section, we briefly discuss some other loss functions that are necessary or beneficial to train our model.\vspace{1mm}

\noindent\textbf{Adversarial Loss}. As Fig.\ref{zhang2} shows, $D$ learns to distinguish real faces from fake ones by minimizing the following adversarial loss~\cite{DBLP:conf/nips/GoodfellowPMXWOCB14}.
\begin{align*}\tag{$17$}\label{eq:17}
\mathcal{L}_{adv}^{D}=&\mathbb{E}_x[(D(x)-1)^2]+\\
&\mathbb{E}_{x_s}[(D(x_s))^2]+\mathbb{E}_{x_f}[(D(x_f))^2]
\end{align*}
where the LSGAN objectives~\cite{DBLP:conf/iccv/MaoLXLWS17} are applied to stabilize the training process and generate faces of higher quality. In contrast, $G$ tries to synthesize fake images to fool $D$ so the adversarial loss of $G$ acts oppositely.
\begin{equation}\tag{$18$}\label{eq:18}
\mathcal{L}_{adv}^{G}=\mathbb{E}_{x_s}[(D(x_s)-1)^2]+\mathbb{E}_{x_f}[(D(x_f)-1)^2]
\end{equation}

\noindent\textbf{KL Loss}. As the random style $m$ is sampled from a prior distribution, the learned makeup code $m_x$ and $m_y$ should also follow the same distribution.
\begin{equation}\tag{$19$}\label{eq:19}
\mathcal{L}_{kl}^{G}=\mathbb{E}_{x,y}[KL(m_x||\mathcal{N}(0,1))+KL(m_y||\mathcal{N}(0,1))]
\end{equation}
where $KL(p||q)=-\int p(z)\log \frac{p(z)}{q(z)}dz$ is the KL divergence.\vspace{1mm}

\noindent\textbf{Total Variation Loss}. To encourage smoothness for the attention mask, we impose the total variation loss~\cite{DBLP:conf/eccv/PumarolaAMSM18} on $M$.
\begin{equation}\tag{$20$}\label{eq:20}
\mathcal{L}_{tv}^G=\mathbb{E}_M[\left\|M_{i+1,j}-M_{i,j}\right\|_1+\left\|M_{i,j+1}-M_{i,j}\right\|_1]
\end{equation}

\noindent\textbf{Full Objective}. By combining the above losses, the full objectives for adversarial learning are defined as follows.
\begin{align*}
\mathcal{L}_D=&\mathcal{L}_{adv}^D\tag{$21$}\label{eq:21}\\
\mathcal{L}_G=&\mathcal{L}_{adv}^G+\lambda_{rec}\mathcal{L}_{rec}^G+\lambda_{per}\mathcal{L}_{per}^G+\\
&\mathcal{L}_{mak}^G+\mathcal{L}_{imr}^G+\lambda_{a}\mathcal{L}_{a}^G+\lambda_{kl}\mathcal{L}_{kl}^G+\lambda_{tv}\mathcal{L}_{tv}^G\tag{$22$}\label{eq:22}
\end{align*}
where $\lambda_{rec},\lambda_{per},\lambda_{face},\lambda_{brow},\lambda_{eye},\lambda_{lip},\lambda_{i},\lambda_{m},\lambda_{a},\lambda_{kl},\lambda_{tv}$ are the weights of different loss terms.

\section{Implementation}

We implement DMT with TensorFlow\footnote{https://www.tensorflow.org/} and conduct all the experiments on a NVIDIA Tesla P100 GPU. We have published an open-source release of our codes as well as the pre-trained model\footnote{https://github.com/Honlan/DMT}.

\begin{figure*}[tb!]
\centering
\includegraphics[width=0.7\textwidth]{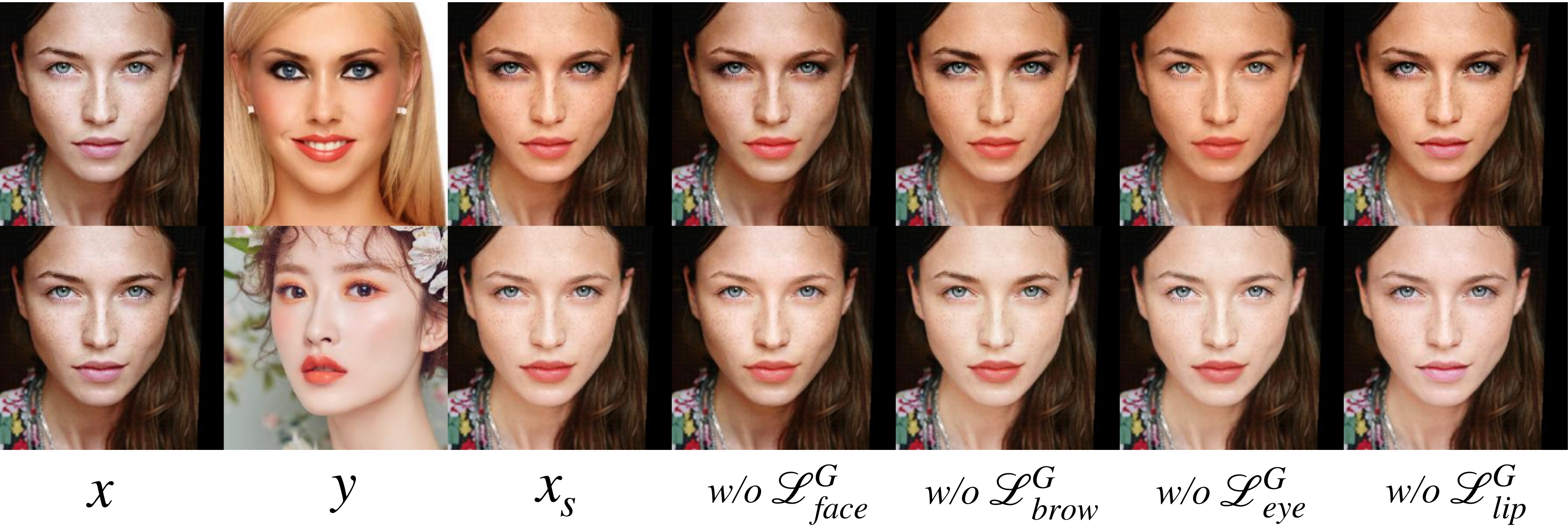}
\caption{Ablation study by removing $\mathcal{L}_{face}^G$, $\mathcal{L}_{brow}^G$, $\mathcal{L}_{eye}^G$, $\mathcal{L}_{lip}^G$ from DMT respectively.}\label{zhang7}
\end{figure*}

In Fig.\ref{zhang5}, we use blocks of different colors to denote different types of neural layers and illustrate the network structures of $E_i$, $E_m$, $G$, $D$ in details. We specify the settings of convolution layers with the attached texts. For example, $k7n64s1$ means a convolution layer with $64$ filters of kernel size $7\times 7$ and stride size $1\times 1$. We apply instance normalization~\cite{DBLP:journals/corr/UlyanovVL16} in $E_i$, adaptive instance normalization~(AdaIN)~\cite{DBLP:conf/iccv/HuangB17} and layer normalization~\cite{DBLP:journals/corr/BaKH16} in $G$, and use relu as the default nonlinearity for $E_i$, $E_m$, $G$. No normalization layers are applied in $E_m$, as they remove the original mean and variance that contain important makeup information. In contrast, $D$ consists of six convolution layers with leaky relu. 

The makeup code of each face image is an $8$-dimensional vector as Fig.\ref{zhang5} shows. In order to blend the information of $i_x$ and $m_y$, we use a multilayer perceptron that takes $m_y$ as input to produce two hidden codes $\gamma$ and $\beta$, which serve as the dynamic mean and variance for the AdaIN layers of $G$. Lastly, $G$ contains two branches to produce the face image $\tilde{x}_s$ with tanh and the attention mask $M$ with sigmoid, which are further combined with $x$ according to Eq.(\ref{eq:15}). 

\section{Experiments}

In this section, we first conduct ablation study to investigate the individual contributions of each component. Then we demonstrate the superiority of our model by comparing against state-of-the-arts. Lastly, we apply our model to perform different scenarios of makeup transfer.

\subsection{Dataset}

We utilize the MT~(Makeup Transfer) dataset released by \cite{DBLP:conf/mm/LiQDLYZL18} to conduct all the experiments, which contains $1,115$ non-makeup and $2,719$ makeup female face images of the resolution $361\times 361$ along with the corresponding parsing masks. We follow the splitting strategy of \cite{DBLP:conf/mm/LiQDLYZL18} by randomly selecting $100$ non-makeup and $250$ makeup images as the test set and use all the other for training.

\subsection{Training}

The training images are resized to $286\times 286$, randomly cropped to $256\times 256$ and horizontally flipped with a probability of $0.5$ for data augmentation. All the neural parameters are initialized with the He initializer~\cite{DBLP:conf/iccv/HeZRS15} and we employ the Adam~\cite{DBLP:journals/corr/KingmaB14} optimizer with $\beta_1=0.5$, $\beta_2=0.999$ for training. 

We set $\lambda_{i}=1$, $\lambda_{m}=1$, $\lambda_{kl}=0.01$, $\lambda_{tv}=0.0001$ following the configurations of \cite{DBLP:conf/eccv/HuangLBK18,DBLP:conf/eccv/LeeTHSY18,DBLP:conf/eccv/PumarolaAMSM18}. As for other weights, we have tried several settings and finally arrive at a proper combination, $\lambda_{rec}=1$, $\lambda_{per}=0.0001$, $\lambda_{face}=\lambda_{brow}=\lambda_{eye}=\lambda_{lip}=50$, $\lambda_{a}=10$, where all the loss terms are sufficiently learned. 

The relu4\_1 layer of VGG16 is used to calculate $\mathcal{L}_{per}^G$. We train DMT for $100$ epochs in all, where the learning rate is fixed as $0.0002$ during the first $50$ epochs and linearly decays to $0$ over the next $50$ epochs. The batch size is set as $1$. For each iteration, we randomly select two training images, makeup or not, then randomly assign them to $x$ and $y$. 

\subsection{Baselines}

We compare our model against the following baselines.

\begin{itemize}
    \item \textbf{DFM}: \textbf{Digital Face Makeup}~\cite{DBLP:conf/cvpr/GuoS09} is an early model based on image processing method.
    \item \textbf{DTN}: \textbf{Deep localized makeup Transfer Network}~\cite{DBLP:conf/ijcai/LiuOQWC16} is an optimization-based model that transfers different cosmetic components separately.
    \item \textbf{BG}: \textbf{BeautyGAN}~\cite{DBLP:conf/mm/LiQDLYZL18} is the state-of-the-art for facial makeup transfer by training a generator with dual inputs and dual outputs.
    \item \textbf{CG}: \textbf{CycleGAN}~\cite{DBLP:journals/corr/ZhuPIE17} can be utilized to achieve facial makeup transfer by treating makeup and non-makeup faces as two domains.
    \item \textbf{ST}: \textbf{Style Transfer}~\cite{DBLP:journals/corr/GatysEB15a} can be utilized to achieve facial makeup transfer by treating the makeup and non-makeup faces as the style and the content.
    \item \textbf{DIA}: \textbf{Deep Image Analogy}~\cite{DBLP:journals/tog/LiaoYYHK17} achieves visual attribute transfer by image analogy to match high-level features extracted from deep neural networks.
\end{itemize}

\begin{figure*}[tb!]
\centering
\includegraphics[width=0.6\textwidth]{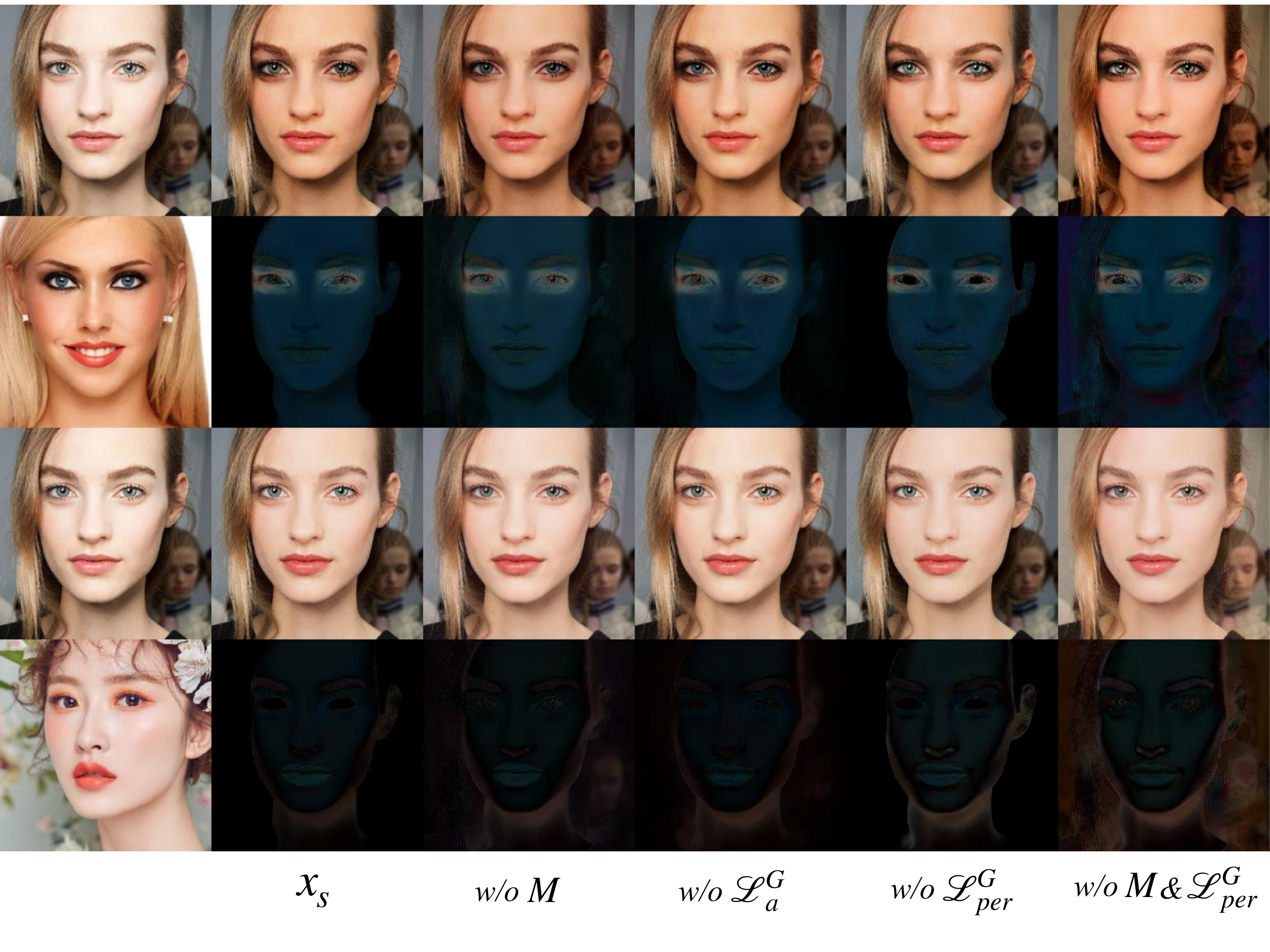}
\caption{Ablation study of the attention mask $M$, the attention loss $\mathcal{L}_a^G$ and the perceptual loss $\mathcal{L}_{per}^G$.}\label{zhang8}
\end{figure*}

\begin{figure*}[tb!]
\centering
\includegraphics[width=0.9\textwidth]{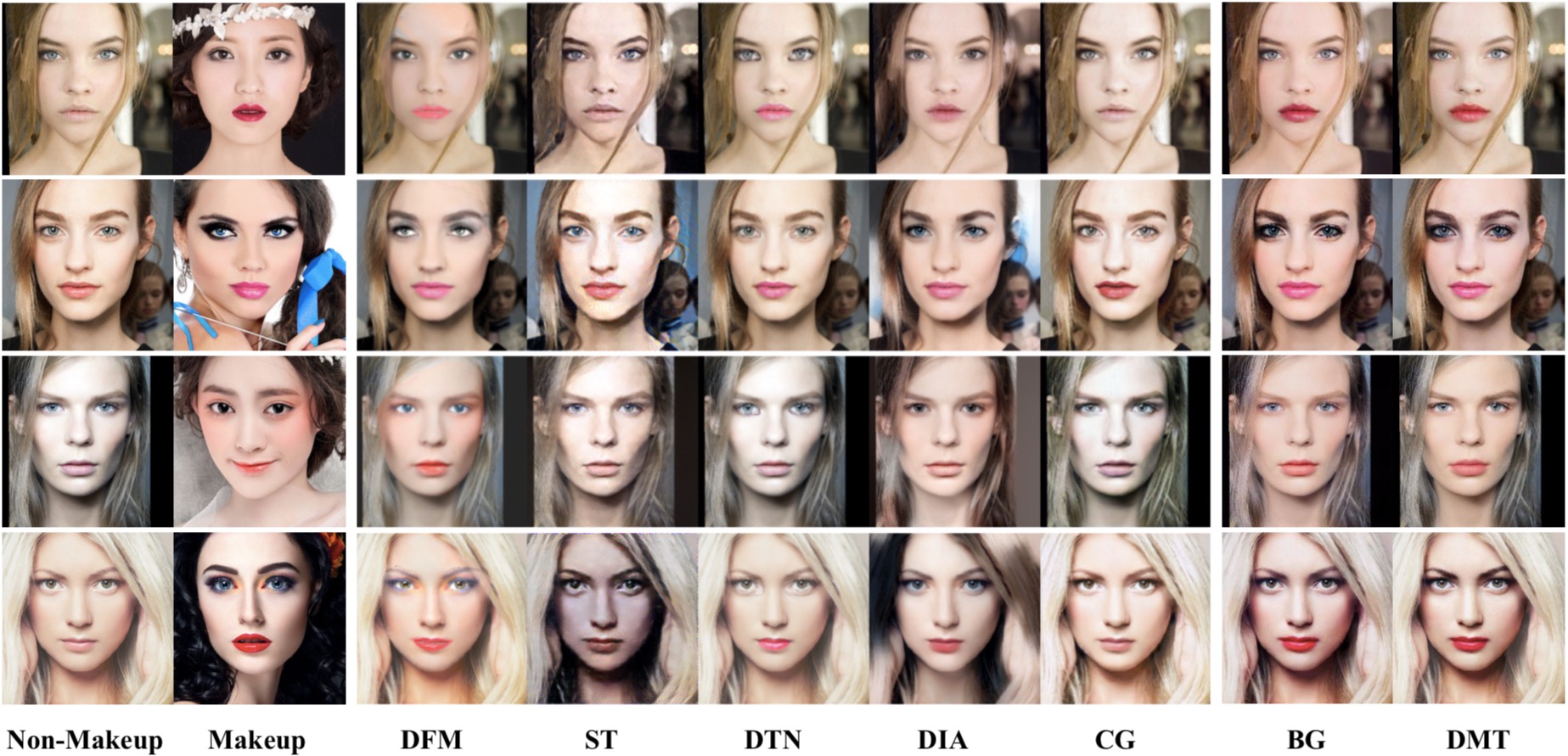}
\caption{Transfer results of DMT against the baselines. DMT can achieve high-quality results and well preserve makeup-unrelated content.}\label{zhang9}
\end{figure*}

\subsection{Ablation Study}

We construct several variants of DMT to investigate the individual contributions of different mechanisms. As Fig.\ref{zhang7} shows, the model fails to accurately add makeup for certain cosmetic components when trained without the corresponding loss terms, including $\mathcal{L}_{face}^G$, $\mathcal{L}_{brow}^G$, $\mathcal{L}_{eye}^G$ and $\mathcal{L}_{lip}^G$. We also investigate the impacts of the attention mask $M$, the attention loss $\mathcal{L}_a^G$ and the perceptual loss $\mathcal{L}_{per}^G$ as Fig.\ref{zhang8} shows, where the \textbf{residual image} $\Delta x$ is employed to visualize the difference between the original non-makeup image $x$ and the transfer result $x_s$.
\begin{equation}\tag{$23$}\label{eq:23}
\Delta x=\left\|x-x_s\right\|_1
\end{equation}

Without $M$~($\mathcal{L}_a^G$ is removed accordingly), we observe that the background is wrongly modified as the residual image shows. After applying $M$ without $\mathcal{L}_a^G$, DMT can learn the makeup-related region in an unsupervised manner, but the background is still slightly altered~(zoom in to see the details). No significant difference is observed without $\mathcal{L}_{per}^G$. However, when both $M$ and $\mathcal{L}_{per}^G$ are removed, the background suffers from obvious changes, which demonstrates that both the attention mask and the perceptual loss contribute to preservation of makeup-unrelated content. 

\begin{figure*}[tb!]
\centering
\includegraphics[width=0.9\textwidth]{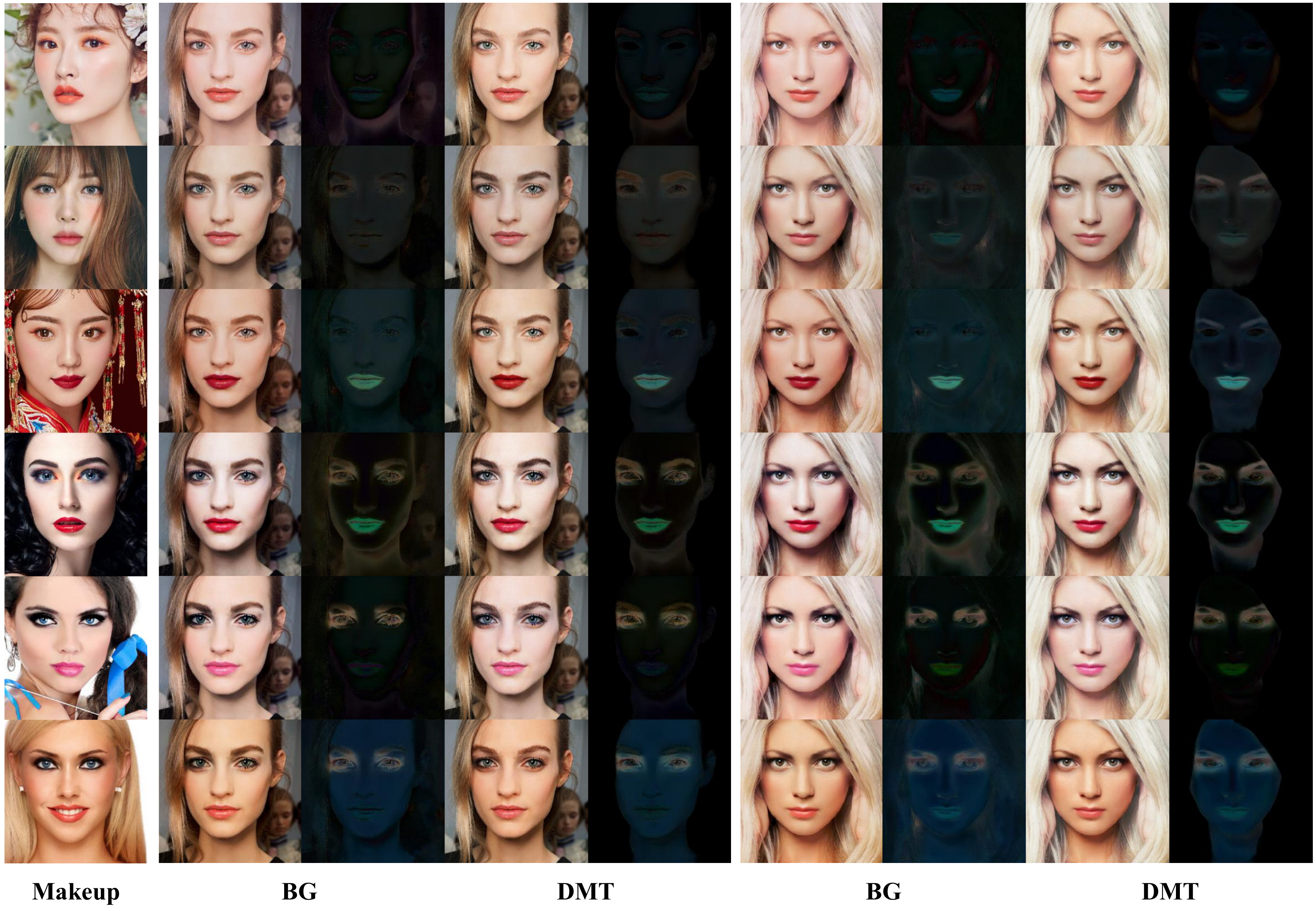}
\caption{Transfer results and residual images of DMT against BG for more makeup styles.}\label{zhang10}
\end{figure*}

\subsection{Qualitative Comparison}

Fig.\ref{zhang9} illustrates the qualitative comparisons of DMT against the baselines on the test set, where the transfer results of DFM, ST, DTN, DIA and CG are provided by \cite{DBLP:conf/mm/LiQDLYZL18}. The results of DFM, DTN and DIA can capture the makeup styles more or less, but all suffer from severe artifacts. ST and CG can generate realistic faces, but fail to add makeup corresponding to the reference images. In contrast, both BG and DMT can produce realistic results of higher quality by properly transferring different cosmetic components. Furthermore, our model is superior to BG by also transferring the eyebrows and better preserving makeup-unrelated content including eyes, hair and background. In subsequent experiments, we mainly compare our model against BG as it outperforms the other five baselines significantly.

We display more comparisons of our model against BG in Fig.\ref{zhang10}. BG can produce visually satisfactory results, but always unavoidably alters makeup-unrelated content according to the residual images. In contrast, DMT can achieve a better tradeoff between makeup transfer and identity preservation by accurately focusing on the crucial cosmetic components.

\subsection{Quantitative Comparison}

We conduct quantitative comparison by human evaluation. Based on the test set, we randomly select $10$ makeup faces for each non-makeup image. As a result, we obtain $1,000$ pairs and conduct pair-wise makeup transfer on them with DMT and BG. The volunteers are instructed to choose the better one according to realism, quality of makeup transfer and preservation of unrelated content. As Table \ref{tab:2} shows, our model outperforms BG by winning $14.6\%$ more votes.

We also compare the reconstruction capability of DMT against BG. For a non-makeup face $x$ and a makeup one $y$, we employ BG to swap the makeup styles for twice to get the reconstructed image $x_r$ and $y_r$. As for DMT, we can simply obtain $x_r$ and $y_r$ according to Eq.(\ref{eq:3}). We perform the above operations on the $1,000$ pairs with DMT and BG respectively to produce two reconstruction sets, both of which contain $2,000$ images. Based on the original images and the reconstructed ones, we leverage the following three metrics, the \textit{Mean Squared Error}~(MSE), the \textit{Peak Signal-to-Noise Ratio}~(PSNR) and the \textit{Structural Similarity Index}~(SSIM)~\cite{DBLP:journals/tip/WangBSS04}, to evaluate the reconstruction capability of DMT and BG. As Table \ref{tab:2} shows, DMT achieves better performances on all the three metrics than BG, which demonstrates that our model can faithfully maintain the one-to-one mappings between face images and identity\,/\,makeup codes.

\begin{table}[tb!]
\centering
\begin{tabular}{c c c c c} \hline
& \textit{human} $\uparrow$ & \textit{MSE} $\downarrow$ & \textit{PSNR} $\uparrow$ & \textit{SSIM} $\uparrow$\\ \hline
\textbf{BG} & $42.7\%$ & 0.00513 & 24.0 & 0.924\\
\textbf{DMT} & $\mathbf{57.3\%}$ & \textbf{0.00028} & \textbf{36.1} & \textbf{0.992}\\ \hline
\end{tabular}
\caption{Quantitative comparisons of DMT against BG, where $\uparrow$ means the higher the better and $\downarrow$ means the lower the better.}\label{tab:2}
\end{table}

\begin{figure*}[tb!]
\centering
\includegraphics[width=0.9\textwidth]{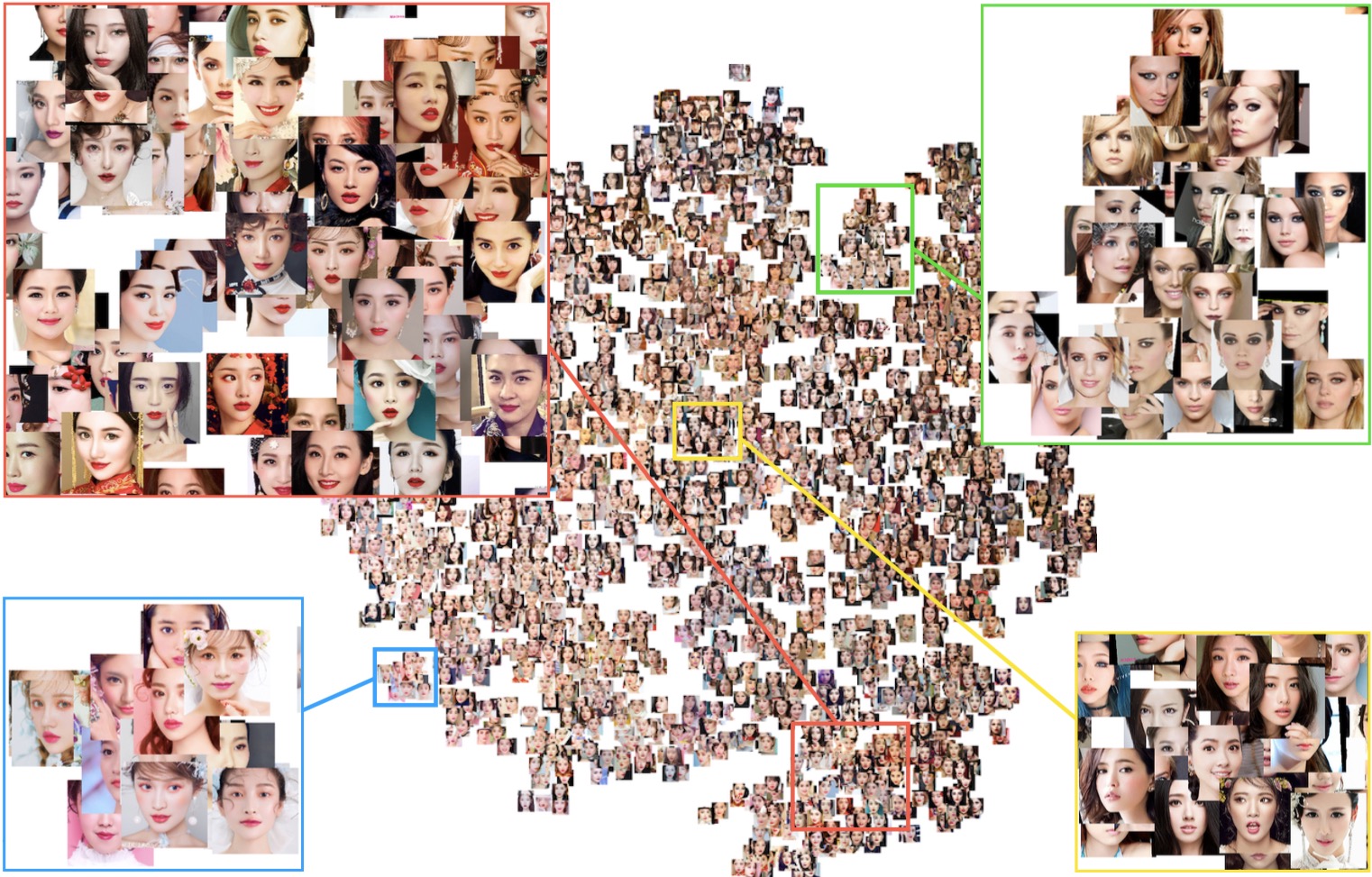}
\caption{Visualization of the learned makeup distribution after dimension reduction.}\label{zhang11}
\end{figure*}

\subsection{Additional Results}

We provide some additional results of DMT on other makeup transfer tasks, which cannot be achieved by BG or other related researches.

To better understand the learned makeup distribution, we calculate the makeup codes of all the makeup faces in the training set and obtain $2,469$ $8$-dimensional vectors. After dimension reduction with t-SNE, we transform each vector into a point in the $2$-D coordinates for visualization. As Fig.\ref{zhang11} shows, faces of similar makeup styles are mapped to closer positions. For example, faces in the green box all belong to smoky-eyes makeup style and those in the red box all wear lipsticks of bright red, which well demonstrate the interpretability of the learned makeup representation.

\begin{figure}[tb!]
\centering
\includegraphics[width=0.45\textwidth]{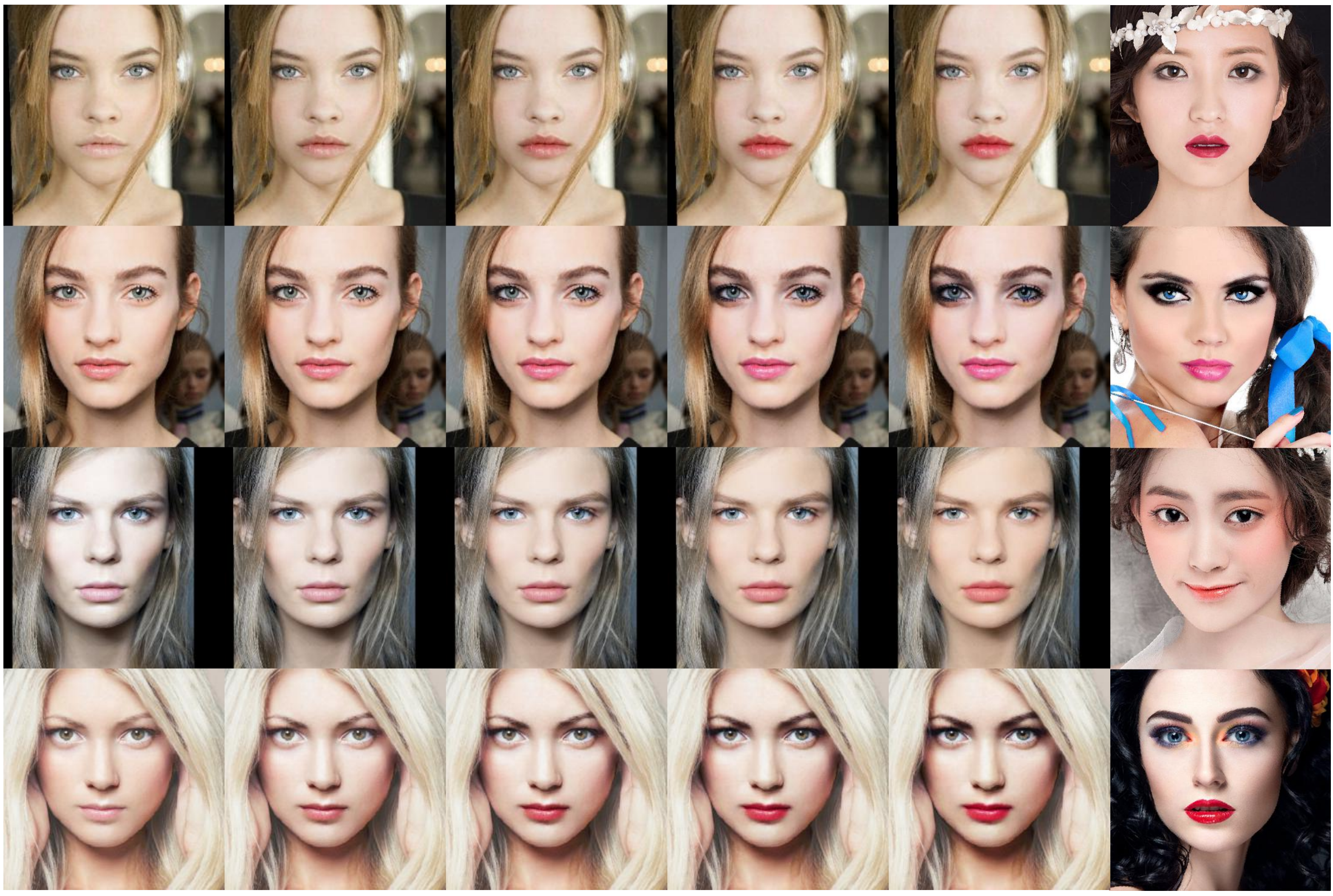}
\caption{Interpolated makeup transfer of DMT by controlling the parameter $\alpha \in[0,1]$.}\label{zhang12}
\end{figure}

\begin{figure}[tb!]
\centering
\includegraphics[width=0.45\textwidth]{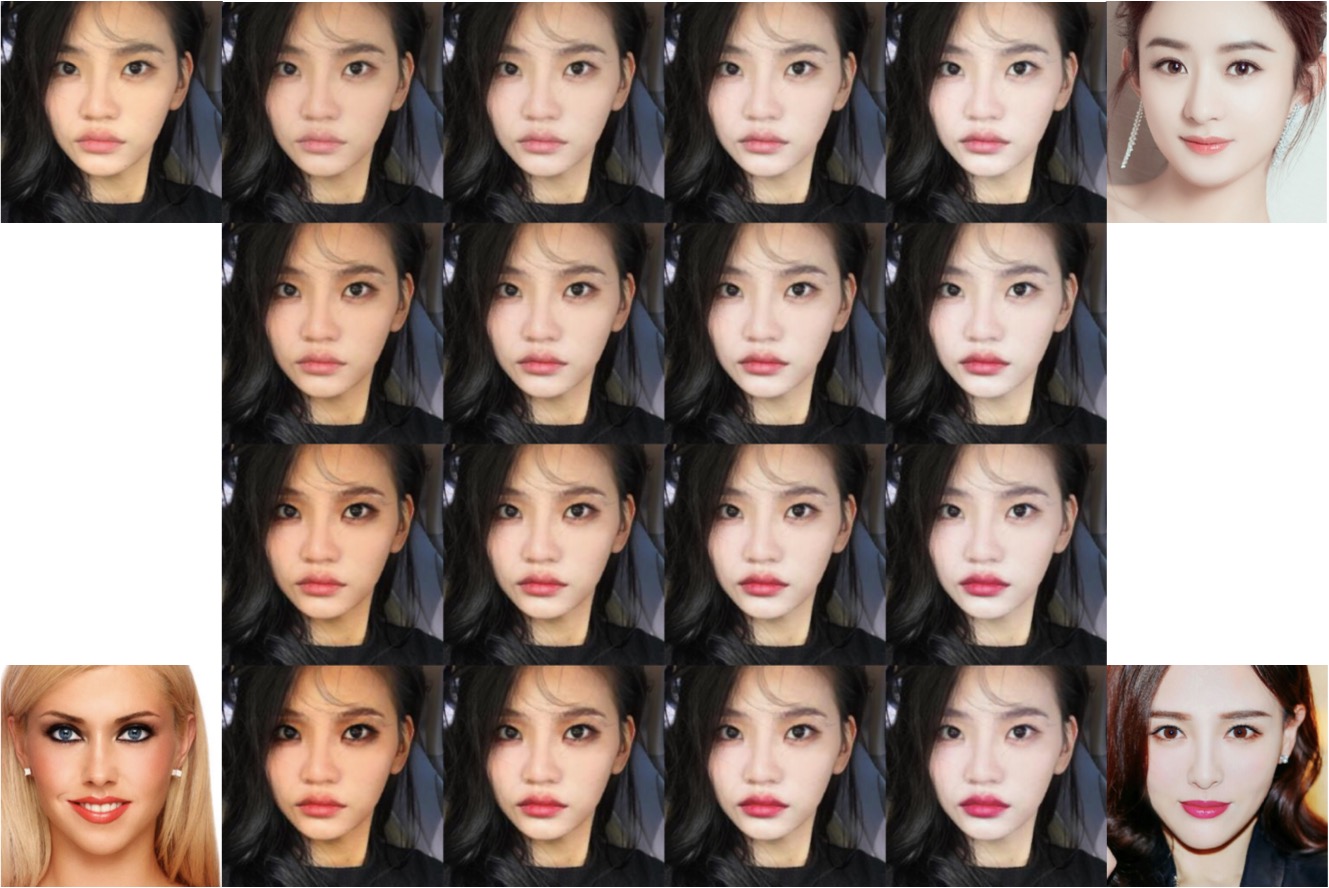}
\caption{hybrid makeup transfer of DMT by combining the makeup codes of multiple faces.}\label{zhang13}
\end{figure}

\begin{figure}[tb!]
\centering
\includegraphics[width=0.45\textwidth]{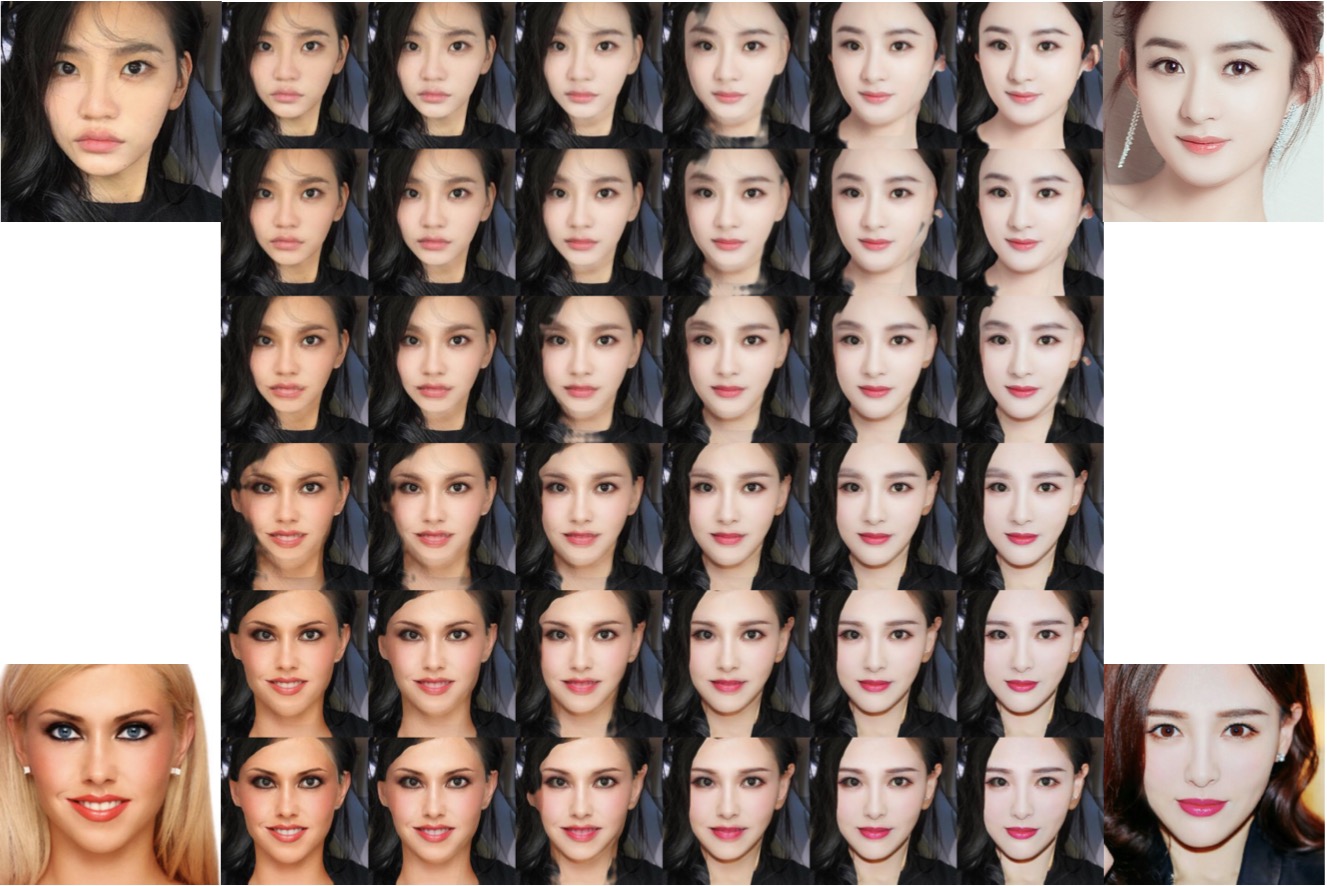}
\caption{Face interpolation of DMT by combining the identity codes and makeup codes of multiple faces.}\label{zhang14}
\end{figure}

\begin{figure}[tb!]
\centering
\includegraphics[width=0.45\textwidth]{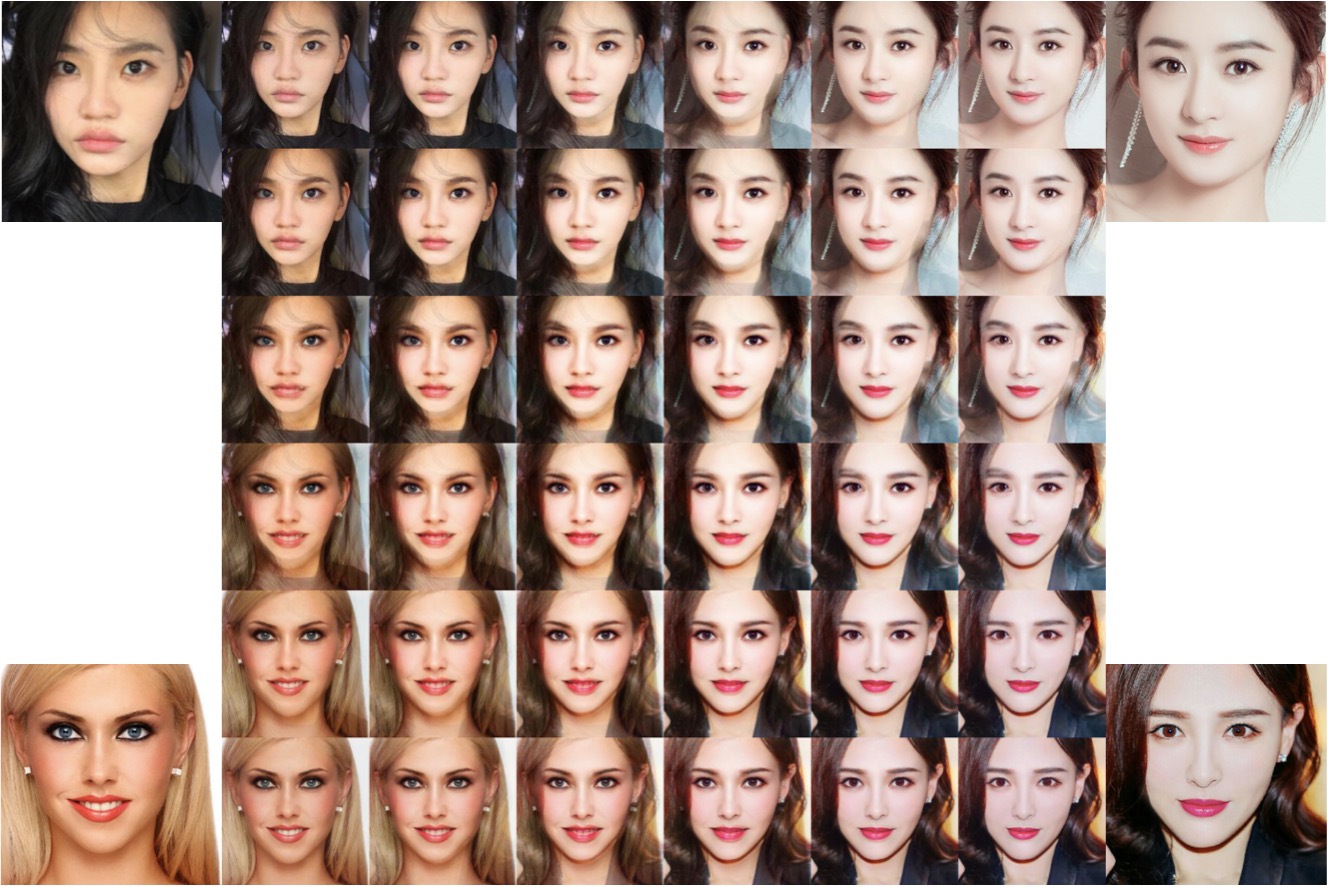}
\caption{Face interpolation of DMT without attention mask by combining the identity codes and makeup codes of multiple faces.}\label{zhang15}
\end{figure}

We employ DMT to conduct interpolated makeup transfer by controlling the parameter $\alpha \in[0,1]$. As Fig.\ref{zhang12} shows, Our model can produce natural results of high quality with increasing strength of makeup style. Based on the disentangled representation, we can also achieve hybrid makeup transfer by blending the makeup codes of multiple reference faces into a non-makeup image as Fig.\ref{zhang13} shows. Another interesting capability of our model is to conduct face interpolation by jointly combining the makeup codes as well as the identity codes. Fig.\ref{zhang14} and Fig.\ref{zhang15} illustrate the face interpolation results of DMT with and without attention mask respectively.

\begin{figure}[tb!]
\centering
\includegraphics[width=0.45\textwidth]{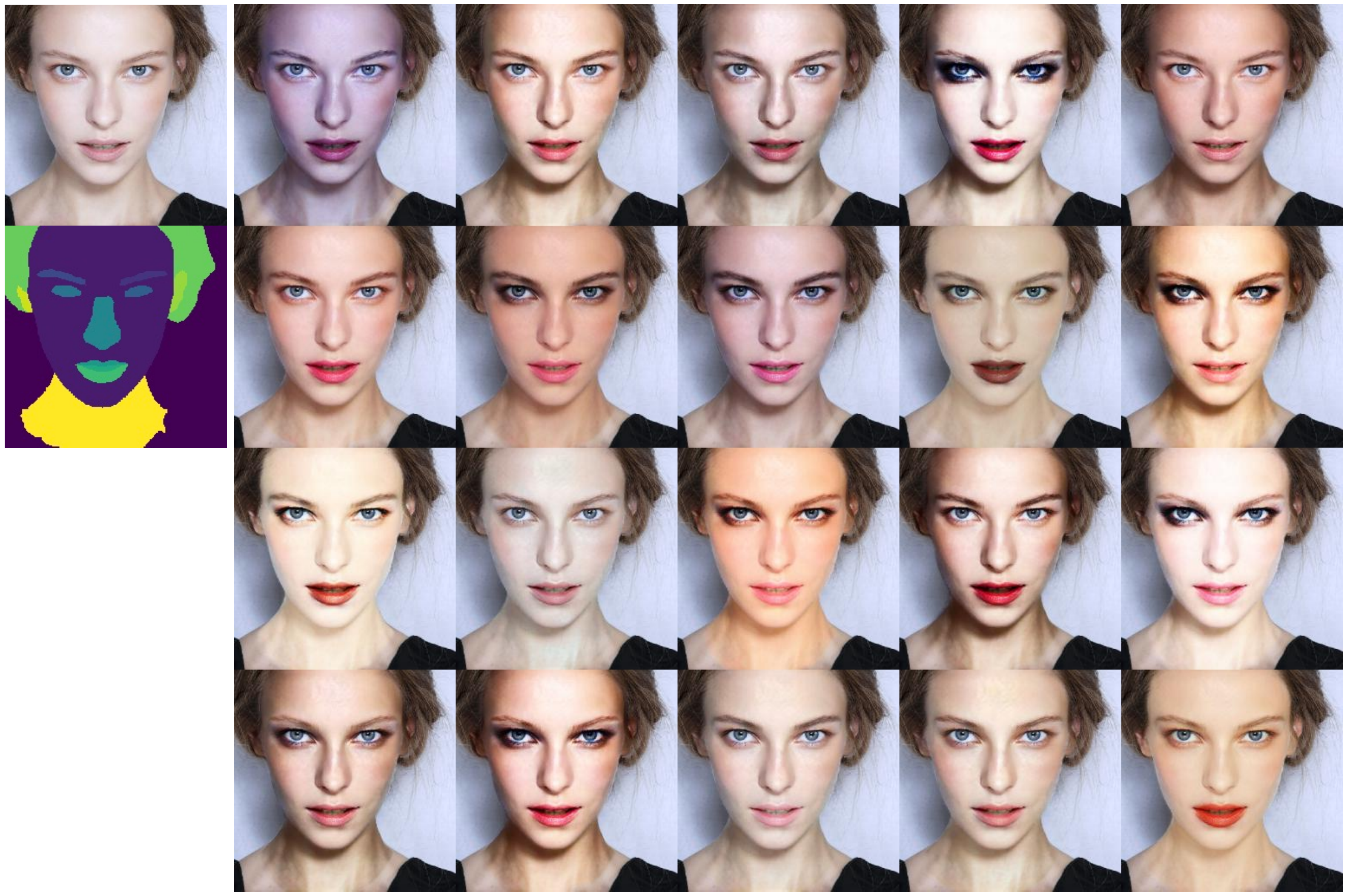}
\caption{Multi-modal makeup transfer of DMT by randomly sampling multiple makeup codes.}\label{zhang16}
\end{figure}

Based on a single non-makeup face, we can achieve multi-modal makeup transfer with DMT by sampling multiple makeup codes from the learned distribution. As Fig.\ref{zhang16} shows, we produce abundant makeup styles diverse in colors of crucial cosmetic components. Most of them look quite appealing and creative, but some may be rare in real life, such as the purple face in the first row. We discover that there exist evident boundary between the neck and the upper body when the color of foundation changes a lot. In fact, this problem is caused by the parsing mask rather than our model, as the semantic part of \textit{neck} does not cover all the visible skin of the upper body~(see the original face and the corresponding parsing mask in Fig.\ref{zhang16}).

\begin{figure}[tb!]
\centering
\includegraphics[width=0.48\textwidth]{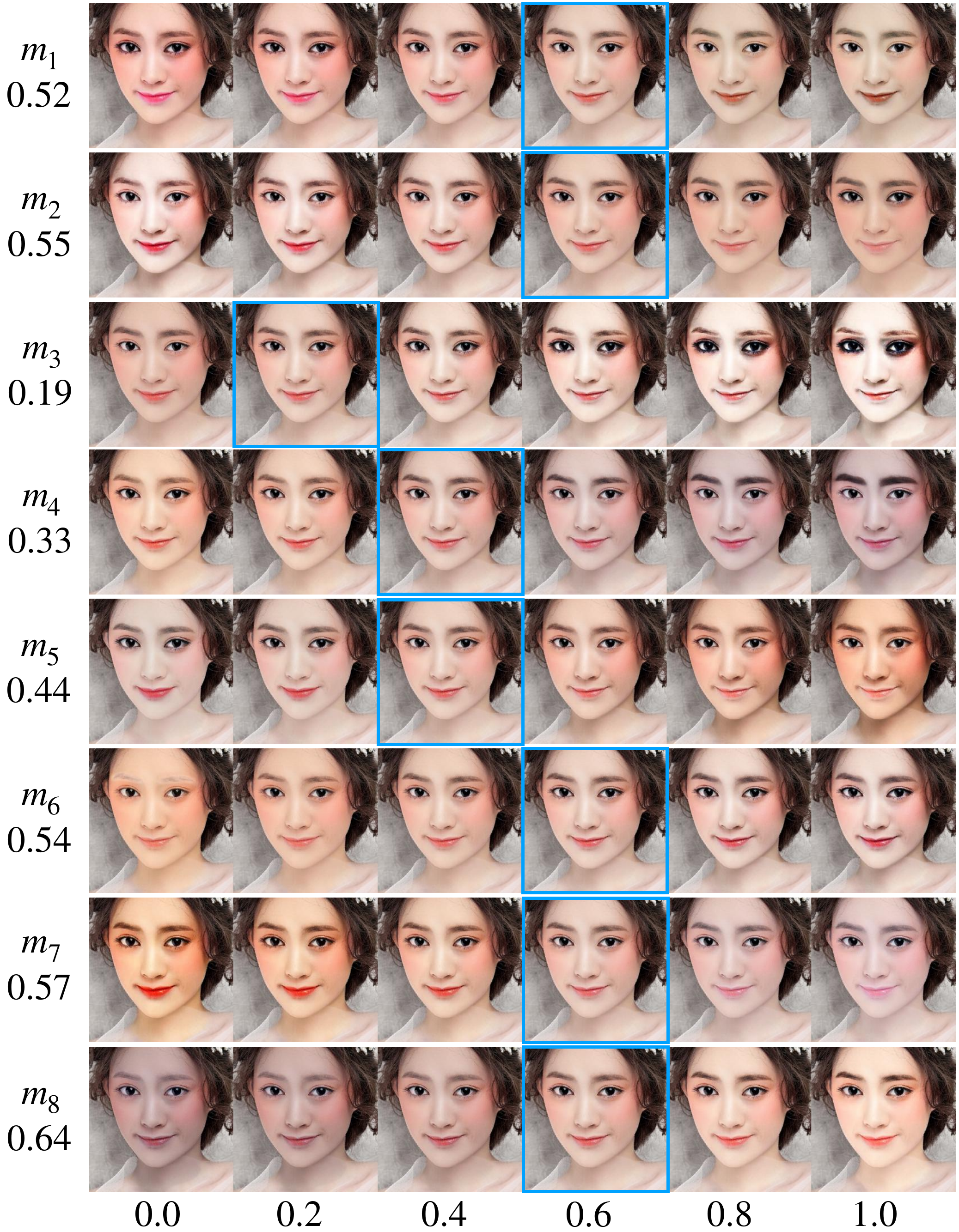}
\caption{Linear interpolation on different dimensions of $m$. In each column, the face in the blue box is the closest one to the input.}\label{zhang17}
\end{figure}

Lastly, we try to interpretate the implications of different dimensions in the makeup code. As Fig.\ref{zhang17} shows, we first calculate the normalized makeup code of a makeup face, $m=(0.52,0.55,0.19,0.33,0.44,0.54,0.57,0.64)$, then adjust the value of each dimension while keeping the others fixed to inspect the corresponding influence. We find that different dimensions are correlated with different makeup styles. For example, increasing the value of $m_3$ results in whiter face and darker eye shadow. It should be noted that the implications of different dimensions are learned in a totally unsupervised manner. If we provide additional annotations like color of lipstick or name of makeup style and correlate them with certain dimensions, the learned makeup code is supposed to be more interpretable and further disentangled. 

\section{Conclusion}

In this paper, we propose DMT~(Disentangled Makeup Transfer), a unified and flexible model to achieve different scenarios of makeup transfer. Our model contains an identity encoder, a makeup encoder and a decoder to learn disentangled representation. We also leverage the attention mask to preserve makeup-unrelated content. Extensive experiments demonstrate that our model can generate better results than state-of-the-arts and perform different scenarios of makeup transfer, which cannot be achieved by related researches. 

\bibliographystyle{named}
\bibliography{ijcai19}

\begin{thebibliography}{}

\bibitem[\protect\citeauthoryear{Ba \bgroup \em et al.\egroup
  }{2016}]{DBLP:journals/corr/BaKH16}
Lei~Jimmy Ba, Ryan Kiros, and Geoffrey~E. Hinton.
\newblock Layer normalization.
\newblock {\em CoRR}, abs/1607.06450, 2016.

\bibitem[\protect\citeauthoryear{Bengio \bgroup \em et al.\egroup
  }{2013}]{DBLP:journals/pami/BengioCV13}
Yoshua Bengio, Aaron~C. Courville, and Pascal Vincent.
\newblock Representation learning: {A} review and new perspectives.
\newblock {\em {IEEE} Trans. Pattern Anal. Mach. Intell.}, 35(8):1798--1828,
  2013.

\bibitem[\protect\citeauthoryear{Chen \bgroup \em et al.\egroup
  }{2018}]{DBLP:conf/eccv/ChenXYT18}
Xinyuan Chen, Chang Xu, Xiaokang Yang, and Dacheng Tao.
\newblock Attention-gan for object transfiguration in wild images.
\newblock In {\em {ECCV}}, pages 167--184, 2018.

\bibitem[\protect\citeauthoryear{Choi \bgroup \em et al.\egroup
  }{2017}]{DBLP:journals/corr/abs-1711-09020}
Yunjey Choi, Min{-}Je Choi, Munyoung Kim, Jung{-}Woo Ha, Sunghun Kim, and
  Jaegul Choo.
\newblock Stargan: Unified generative adversarial networks for multi-domain
  image-to-image translation.
\newblock {\em CoRR}, abs/1711.09020, 2017.

\bibitem[\protect\citeauthoryear{Gatys \bgroup \em et al.\egroup
  }{2015}]{DBLP:journals/corr/GatysEB15a}
Leon~A. Gatys, Alexander~S. Ecker, and Matthias Bethge.
\newblock A neural algorithm of artistic style.
\newblock {\em CoRR}, abs/1508.06576, 2015.

\bibitem[\protect\citeauthoryear{Goodfellow \bgroup \em et al.\egroup
  }{2014}]{DBLP:conf/nips/GoodfellowPMXWOCB14}
Ian~J. Goodfellow, Jean Pouget{-}Abadie, Mehdi Mirza, Bing Xu, David
  Warde{-}Farley, Sherjil Ozair, Aaron~C. Courville, and Yoshua Bengio.
\newblock Generative adversarial nets.
\newblock In {\em {NeurIPS}}, pages 2672--2680, 2014.

\bibitem[\protect\citeauthoryear{Gulrajani \bgroup \em et al.\egroup
  }{2017}]{DBLP:conf/nips/GulrajaniAADC17}
Ishaan Gulrajani, Faruk Ahmed, Mart{\'{\i}}n Arjovsky, Vincent Dumoulin, and
  Aaron~C. Courville.
\newblock Improved training of wasserstein gans.
\newblock In {\em {NeurIPS}}, pages 5769--5779, 2017.

\bibitem[\protect\citeauthoryear{Guo and Sim}{2009}]{DBLP:conf/cvpr/GuoS09}
Dong Guo and Terence Sim.
\newblock Digital face makeup by example.
\newblock In {\em {CVPR}}, pages 73--79, 2009.

\bibitem[\protect\citeauthoryear{He \bgroup \em et al.\egroup
  }{2015}]{DBLP:conf/iccv/HeZRS15}
Kaiming He, Xiangyu Zhang, Shaoqing Ren, and Jian Sun.
\newblock Delving deep into rectifiers: Surpassing human-level performance on
  imagenet classification.
\newblock In {\em {ICCV}}, pages 1026--1034, 2015.

\bibitem[\protect\citeauthoryear{Huang and
  Belongie}{2017}]{DBLP:conf/iccv/HuangB17}
Xun Huang and Serge~J. Belongie.
\newblock Arbitrary style transfer in real-time with adaptive instance
  normalization.
\newblock In {\em {ICCV}}, pages 1510--1519, 2017.

\bibitem[\protect\citeauthoryear{Huang \bgroup \em et al.\egroup
  }{2018}]{DBLP:conf/eccv/HuangLBK18}
Xun Huang, Ming{-}Yu Liu, Serge~J. Belongie, and Jan Kautz.
\newblock Multimodal unsupervised image-to-image translation.
\newblock In {\em {ECCV}}, pages 179--196, 2018.

\bibitem[\protect\citeauthoryear{Isola \bgroup \em et al.\egroup
  }{2017}]{DBLP:conf/cvpr/IsolaZZE17}
Phillip Isola, Jun{-}Yan Zhu, Tinghui Zhou, and Alexei~A. Efros.
\newblock Image-to-image translation with conditional adversarial networks.
\newblock In {\em {CVPR}}, pages 5967--5976, 2017.

\bibitem[\protect\citeauthoryear{Johnson \bgroup \em et al.\egroup
  }{2016}]{DBLP:conf/eccv/JohnsonAF16}
Justin Johnson, Alexandre Alahi, and Li~Fei{-}Fei.
\newblock Perceptual losses for real-time style transfer and super-resolution.
\newblock In {\em {ECCV}}, pages 694--711, 2016.

\bibitem[\protect\citeauthoryear{Kim \bgroup \em et al.\egroup
  }{2017}]{DBLP:conf/icml/KimCKLK17}
Taeksoo Kim, Moonsu Cha, Hyunsoo Kim, Jung~Kwon Lee, and Jiwon Kim.
\newblock Learning to discover cross-domain relations with generative
  adversarial networks.
\newblock In {\em {ICML}}, pages 1857--1865, 2017.

\bibitem[\protect\citeauthoryear{Kingma and
  Ba}{2014}]{DBLP:journals/corr/KingmaB14}
Diederik~P. Kingma and Jimmy Ba.
\newblock Adam: {A} method for stochastic optimization.
\newblock {\em CoRR}, abs/1412.6980, 2014.

\bibitem[\protect\citeauthoryear{Ledig \bgroup \em et al.\egroup
  }{2017}]{DBLP:conf/cvpr/LedigTHCCAATTWS17}
Christian Ledig, Lucas Theis, Ferenc Huszar, Jose Caballero, Andrew Cunningham,
  Alejandro Acosta, Andrew~P. Aitken, Alykhan Tejani, Johannes Totz, Zehan
  Wang, and Wenzhe Shi.
\newblock Photo-realistic single image super-resolution using a generative
  adversarial network.
\newblock In {\em {CVPR}}, pages 105--114, 2017.

\bibitem[\protect\citeauthoryear{Lee \bgroup \em et al.\egroup
  }{2018}]{DBLP:conf/eccv/LeeTHSY18}
Hsin{-}Ying Lee, Hung{-}Yu Tseng, Jia{-}Bin Huang, Maneesh Singh, and
  Ming{-}Hsuan Yang.
\newblock Diverse image-to-image translation via disentangled representations.
\newblock In {\em {ECCV}}, pages 36--52, 2018.

\bibitem[\protect\citeauthoryear{Lee \bgroup \em et al.\egroup
  }{2019}]{CelebAMask-HQ}
Cheng-Han Lee, Ziwei Liu, Lingyun Wu, and Ping Luo.
\newblock Maskgan: Towards diverse and interactive facial image manipulation.
\newblock {\em Technical Report}, 2019.

\bibitem[\protect\citeauthoryear{Li \bgroup \em et al.\egroup
  }{2015}]{DBLP:conf/cvpr/LiZL15}
Chen Li, Kun Zhou, and Stephen Lin.
\newblock Simulating makeup through physics-based manipulation of intrinsic
  image layers.
\newblock In {\em {CVPR}}, pages 4621--4629, 2015.

\bibitem[\protect\citeauthoryear{Li \bgroup \em et al.\egroup
  }{2018}]{DBLP:conf/mm/LiQDLYZL18}
Tingting Li, Ruihe Qian, Chao Dong, Si~Liu, Qiong Yan, Wenwu Zhu, and Liang
  Lin.
\newblock Beautygan: Instance-level facial makeup transfer with deep generative
  adversarial network.
\newblock In {\em {ACM MM}}, pages 645--653, 2018.

\bibitem[\protect\citeauthoryear{Liao \bgroup \em et al.\egroup
  }{2017}]{DBLP:journals/tog/LiaoYYHK17}
Jing Liao, Yuan Yao, Lu~Yuan, Gang Hua, and Sing~Bing Kang.
\newblock Visual attribute transfer through deep image analogy.
\newblock {\em {ACM} Trans. Graph.}, 36(4):120:1--120:15, 2017.

\bibitem[\protect\citeauthoryear{Liu \bgroup \em et al.\egroup
  }{2016}]{DBLP:conf/ijcai/LiuOQWC16}
Si~Liu, Xinyu Ou, Ruihe Qian, Wei Wang, and Xiaochun Cao.
\newblock Makeup like a superstar: Deep localized makeup transfer network.
\newblock In {\em {IJCAI}}, pages 2568--2575, 2016.

\bibitem[\protect\citeauthoryear{Liu \bgroup \em et al.\egroup
  }{2018}]{DBLP:conf/eccv/LiuRSWTC18}
Guilin Liu, Fitsum~A. Reda, Kevin~J. Shih, Ting{-}Chun Wang, Andrew Tao, and
  Bryan Catanzaro.
\newblock Image inpainting for irregular holes using partial convolutions.
\newblock In {\em {ECCV}}, pages 89--105, 2018.

\bibitem[\protect\citeauthoryear{Ma \bgroup \em et al.\egroup
  }{2018}]{DBLP:conf/cvpr/MaSGGSF18}
Liqian Ma, Qianru Sun, Stamatios Georgoulis, Luc~Van Gool, Bernt Schiele, and
  Mario Fritz.
\newblock Disentangled person image generation.
\newblock In {\em {CVPR}}, pages 99--108, 2018.

\bibitem[\protect\citeauthoryear{Mao \bgroup \em et al.\egroup
  }{2017}]{DBLP:conf/iccv/MaoLXLWS17}
Xudong Mao, Qing Li, Haoran Xie, Raymond Y.~K. Lau, Zhen Wang, and Stephen~Paul
  Smolley.
\newblock Least squares generative adversarial networks.
\newblock In {\em {ICCV}}, pages 2813--2821, 2017.

\bibitem[\protect\citeauthoryear{Mejjati \bgroup \em et al.\egroup
  }{2018}]{DBLP:conf/nips/MejjatiRTCK18}
Youssef~Alami Mejjati, Christian Richardt, James Tompkin, Darren Cosker, and
  Kwang~In Kim.
\newblock Unsupervised attention-guided image-to-image translation.
\newblock In {\em {NeurIPS}}, pages 3697--3707, 2018.

\bibitem[\protect\citeauthoryear{Pumarola \bgroup \em et al.\egroup
  }{2018}]{DBLP:conf/eccv/PumarolaAMSM18}
Albert Pumarola, Antonio Agudo, Aleix~M. Martinez, Alberto Sanfeliu, and
  Francesc Moreno{-}Noguer.
\newblock Ganimation: Anatomically-aware facial animation from a single image.
\newblock In {\em {ECCV}}, pages 835--851, 2018.

\bibitem[\protect\citeauthoryear{Radford \bgroup \em et al.\egroup
  }{2015}]{DBLP:journals/corr/RadfordMC15}
Alec Radford, Luke Metz, and Soumith Chintala.
\newblock Unsupervised representation learning with deep convolutional
  generative adversarial networks.
\newblock {\em CoRR}, abs/1511.06434, 2015.

\bibitem[\protect\citeauthoryear{Russakovsky \bgroup \em et al.\egroup
  }{2015}]{DBLP:journals/ijcv/RussakovskyDSKS15}
Olga Russakovsky, Jia Deng, Hao Su, Jonathan Krause, Sanjeev Satheesh, Sean Ma,
  Zhiheng Huang, Andrej Karpathy, Aditya Khosla, Michael~S. Bernstein,
  Alexander~C. Berg, and Fei{-}Fei Li.
\newblock Imagenet large scale visual recognition challenge.
\newblock {\em {IJCV}}, 115(3):211--252, 2015.

\bibitem[\protect\citeauthoryear{Simonyan and
  Zisserman}{2015}]{DBLP:journals/corr/SimonyanZ14a}
Karen Simonyan and Andrew Zisserman.
\newblock Very deep convolutional networks for large-scale image recognition.
\newblock In {\em {ICLR}}, 2015.

\bibitem[\protect\citeauthoryear{Smith \bgroup \em et al.\egroup
  }{2013}]{DBLP:conf/cvpr/SmithZBLY13}
Brandon~M. Smith, Li~Zhang, Jonathan Brandt, Zhe Lin, and Jianchao Yang.
\newblock Exemplar-based face parsing.
\newblock In {\em {CVPR}}, pages 3484--3491, 2013.

\bibitem[\protect\citeauthoryear{Tong \bgroup \em et al.\egroup
  }{2007}]{DBLP:conf/pg/TongTBX07}
Wai{-}Shun Tong, Chi{-}Keung Tang, Michael~S. Brown, and Ying{-}Qing Xu.
\newblock Example-based cosmetic transfer.
\newblock In {\em {PCCGA}}, pages 211--218, 2007.

\bibitem[\protect\citeauthoryear{Ulyanov \bgroup \em et al.\egroup
  }{2016}]{DBLP:journals/corr/UlyanovVL16}
Dmitry Ulyanov, Andrea Vedaldi, and Victor~S. Lempitsky.
\newblock Instance normalization: The missing ingredient for fast stylization.
\newblock {\em CoRR}, abs/1607.08022, 2016.

\bibitem[\protect\citeauthoryear{Wang \bgroup \em et al.\egroup
  }{2004}]{DBLP:journals/tip/WangBSS04}
Zhou Wang, Alan~C. Bovik, Hamid~R. Sheikh, and Eero~P. Simoncelli.
\newblock Image quality assessment: from error visibility to structural
  similarity.
\newblock {\em {IEEE} Trans. Image Processing}, 13(4):600--612, 2004.

\bibitem[\protect\citeauthoryear{Yang \bgroup \em et al.\egroup
  }{2018}]{DBLP:journals/corr/abs-1806-06195}
Chao Yang, Taehwan Kim, Ruizhe Wang, Hao Peng, and C.{-}C.~Jay Kuo.
\newblock Show, attend and translate: Unsupervised image translation with
  self-regularization and attention.
\newblock {\em CoRR}, abs/1806.06195, 2018.

\bibitem[\protect\citeauthoryear{Yi \bgroup \em et al.\egroup
  }{2017}]{DBLP:conf/iccv/YiZTG17}
Zili Yi, Hao~(Richard) Zhang, Ping Tan, and Minglun Gong.
\newblock Dualgan: Unsupervised dual learning for image-to-image translation.
\newblock In {\em {ICCV}}, pages 2868--2876, 2017.

\bibitem[\protect\citeauthoryear{Yu \bgroup \em et al.\egroup
  }{2018}]{DBLP:conf/eccv/YuWPGYS18}
Changqian Yu, Jingbo Wang, Chao Peng, Changxin Gao, Gang Yu, and Nong Sang.
\newblock Bisenet: Bilateral segmentation network for real-time semantic
  segmentation.
\newblock In {\em {ECCV}}, pages 334--349, 2018.

\bibitem[\protect\citeauthoryear{Zhang \bgroup \em et al.\egroup
  }{2017}]{DBLP:conf/iccv/ZhangXL17}
Han Zhang, Tao Xu, and Hongsheng Li.
\newblock Stackgan: Text to photo-realistic image synthesis with stacked
  generative adversarial networks.
\newblock In {\em {ICCV}}, pages 5908--5916, 2017.

\bibitem[\protect\citeauthoryear{Zhang \bgroup \em et al.\egroup
  }{2018}]{DBLP:conf/eccv/ZhangKSC18}
Gang Zhang, Meina Kan, Shiguang Shan, and Xilin Chen.
\newblock Generative adversarial network with spatial attention for face
  attribute editing.
\newblock In {\em {ECCV}}, pages 422--437, 2018.

\bibitem[\protect\citeauthoryear{Zhao \bgroup \em et al.\egroup
  }{2017}]{DBLP:conf/cvpr/ZhaoSQWJ17}
Hengshuang Zhao, Jianping Shi, Xiaojuan Qi, Xiaogang Wang, and Jiaya Jia.
\newblock Pyramid scene parsing network.
\newblock In {\em {CVPR}}, pages 6230--6239, 2017.

\bibitem[\protect\citeauthoryear{Zhu \bgroup \em et al.\egroup
  }{2017}]{DBLP:journals/corr/ZhuPIE17}
Jun{-}Yan Zhu, Taesung Park, Phillip Isola, and Alexei~A. Efros.
\newblock Unpaired image-to-image translation using cycle-consistent
  adversarial networks.
\newblock {\em CoRR}, abs/1703.10593, 2017.

\end{thebibliography}

\end{document}